\documentclass[twoside,english]{elsarticle}
\usepackage[T1]{fontenc}
\usepackage[latin9]{inputenc}
\usepackage{float}
\usepackage{amsmath}
\usepackage{amsthm}
\usepackage{amssymb}
\usepackage{graphicx}

\makeatletter
\theoremstyle{plain}
\newtheorem{thm}{\protect\theoremname}
\theoremstyle{definition}
\newtheorem{defn}[thm]{\protect\definitionname}

\journal{Example: Nuclear Physics B}

\makeatother

\usepackage{babel}
\providecommand{\definitionname}{Definition}
\providecommand{\theoremname}{Theorem}

\begin{document}

\begin{frontmatter}{}

\title{Towards secure judgments aggregation in AHP}

\author[kis]{Konrad Ku\l akowski\corref{cor1}}

\ead{konrad.kulakowski@agh.edu.pl}

\author[wms]{Jacek~Szybowski}

\ead{jacek.szybowski@agh.edu.pl}

\author[opf]{Jiri~Mazurek}

\ead{mazurek@opf.slu.cz}

\author[kis]{Sebastian~Ernst}

\ead{sebastian.ernst@agh.edu.pl}

\cortext[cor1]{Corresponding author}

\address[kis]{AGH, University of Science and Technology, Department of Applied
Computer Science }

\address[opf]{Department of Informatics and Mathematics, Silesian University in
Opava, School of Business Administration in Karvina, Univerzitni namesti
1934/3, Czech Republic}

\address[wms]{AGH, University of Science and Technology, Faculty of Applied Mathematics}
\begin{abstract}
In decision-making methods, it is common to assume that the experts
are honest and professional. However, this is not the case when one
or more experts in the group decision making framework, such as the
group analytic hierarchy process (GAHP), try to manipulate results
in their favor. The aim of this paper is to introduce two heuristics
in the GAHP, setting allowing to detect the manipulators and minimize
their effect on the group consensus by diminishing their weights.
The first heuristic is based on the assumption that manipulators will
provide judgments which can be considered outliers with respect to
those of the rest of the experts in the group. The second heuristic
assumes that dishonest judgments are less consistent than the average
consistency of the group. Both approaches are illustrated with numerical
examples and simulations.
\end{abstract}
\begin{keyword}
pairwise comparisons \sep manipulation \sep group decision-making
\sep AHP \sep Analytic Hierarchy Process 
\end{keyword}

\end{frontmatter}{}

\section{Introduction\label{sec:Introduction}}

Group decision making refers to the situations where the problem of
selecting the best alternative (option, solution, etc.) is handled
collectively by a set of individuals, preferably experts in the field.
Usually, such processes involve complex problems, too difficult to
be handled by an individual, or they deal with actions legally required
to be made by a collective. These situations include government meetings,
various board negotiations, policy making, business dealings, complex
laboratory experiments, elections, jury trials, reaching consensus
in social networks, and many others. The fundamentals of group decision
making can be found e.g. in \citep{Castelan1993iagd,Dong2016cbig,GarciaZamora2022lsgd,Hwang1987gdmu,Kilgour2021hogd,Lootsma1999gdm,Lootsma1999mcdm,Saaty1989gdma}.

One of the problems associated with group decision making (GDM) is
that it is susceptible to manipulation if one or more experts try
to influence the group outcome in their favor by, for example, providing
dishonest judgments. The manipulation, especially in the political
context, can be traced back at least to the ancient societies of Greece
and Rome, see e.g. \citep{Connor1987tfap}. More recently, a description
of political manipulation in a GDM setting can be found, for example,
in the work of Maoz \citep{Maoz1990ftni}, who investigated examples
of U.S. and Israeli foreign policy choices under crisis conditions,
or Hoyt \citep{Hoyt1997tpmo}, who examined the American decision
process during the Iranian revolution. Analysis of manipulation in
selected voting methods can be found e.g. in \citep{Brandt2016hocs,Gardenfors1976mosc,Gibbard1973movs,Macintyre1993mumd,Smith1999maoc,Taylor2005scat}.
Further on, Faliszewski et al. \citep{Faliszewski2009lacv,Faliszewski2010uctp}
proposed different approaches of manipulation protection in the context
of elections.

Another area of group decision making vulnerable to manipulation are
social networks. An approach to prevent weight manipulation by minimum
adjustment and maximum entropy method in social network group decision
making can be found in \citep{Sun2022aatp}. Similarly, Wu et al.
\citep{Wu2021aofm} introduced a novel framework to prevent manipulative
behaviors during consensus-reaching processes in social network group
decision making. They considered two means of manipulation \textendash{}
individual manipulation, where each expert manipulates his/her own
behavior to achieve higher importance (weight); and group manipulation,
where a group of experts forces inconsistent experts to adopt specific
recommendation advices \textendash{} and investigated models to counteract
both of them. Manipulation in multiple-criteria group decision making
attracted the attention of several recent studies. Dong et al. \citep{Dong2021cras}
presented a new strategic manipulation called trust relationship manipulation
and discussed clique-based strategies to manipulate trust relationships
to obtain the desired ranking of the alternatives. Hnatiienko \citep{Hnatiienko2019cmim}
studied the problem of manipulating the choice of decision options
in peer review processes and proposed a classification of selection
manipulation problems in experts\textquoteright{} evaluation. Lev
and Lewenberg \citep{Lev2019rgmi} investigated cases when agents
may wish to re-draw the organizational chart of a company, or markets
(which is called \textquoteleft reverse gerrymandering\textquoteright ),
to maximize their influence across the company\textquoteright s sub-units,
or to allocate resources to the desired areas. Yager \citep{Yager2001pspm,Yager2002dasm}
studied methods of strategic manipulation of preferential data. He
proposed modification of the preference aggregation function in such
a way that the attempts of individual agents to manipulate the data
are penalized. Dong et al. \citep{Dong2018swmi} defined the concept
of the ranking range of an alternative in multiple-attribute decision
making and proposed a series of mixed binary linear programming models
to show the process of designing a strategic attribute weight vector.
Moreover, the authors studied the conditions to manipulate a strategic
attribute weight based on the ranking range and the proposed model.
Sasaki \citep{Sasaki2023smig} discussed the issue of strategic manipulation
in the context of group decision making with pairwise comparisons.
The author considered a scenario of group decision-making situations
formulated as strategic games and his theoretical results show truthful
judgments (pairwise comparisons) can be a dominant strategy only in
very few situations.

Apart from the last study, the problem of manipulation in pairwise
comparison methods has not as yet been studied thoroughly. This paper
aims to fill the aforementioned gap and focuses on group decision
making in the analytic hierarchy process (GAHP), along with the problem
of possible manipulation of its outcome. In the GAHP setting, a group
of experts provides pairwise comparisons of alternatives under consideration
with the aim of selecting the best alternative, see e.g. Dong and
Saaty \citep{Dong2014aahp}, Ramanathan and Ganesh \citep{Ramanathan1994gpam},
or Saaty \citep{Saaty1989gdma}. The aim of the paper is to introduce
two heuristics in the GAHP setting to detect the manipulators and
minimize their effect on the group consensus by minimizing their weights.
The first heuristic is based on the assumption that manipulators will
provide judgments which can be considered outliers with respect to
judgments of the rest of the experts in the group. The second one
assumes that dishonest judgments are less consistent than the average
consistency of the group. Both approaches are illustrated with numerical
examples and simulations.

The paper is composed of five sections. The \emph{Introduction} (Sec.
\ref{sec:Introduction}) and \emph{Preliminaries} (Sec. \ref{sec:Preliminaries})
aim to introduce the reader to the literature on the subject and recall
the necessary concepts and definitions of the quantitative and qualitative
pairwise comparisons method. The following section, \emph{Inconsistency
Driven Pairwise Ranking Aggregation} (Sec. \ref{sec:Inconsistency-Driven-Pairwise})\emph{,}
identifies the problem of ranking manipulation and introduces the
proposed robust methods of aggregating results coming from various
experts. Sec. \ref{sec:Montecarlo-experiments} contains two Montecarlo
experiments allowing to assess the effectiveness of the proposed methods.
The paper ends with (Sec. \ref{sec:Summary}), containing a short
summary of the achieved results.

\noindent 

\section{Preliminaries\label{sec:Preliminaries}}

\subsection{Pairwise comparisons\label{subsec:Pairwise-comparisons}}

Comparing alternatives in pairs serves as the basis of many decision-making
methods, including AHP, BWM, HRE, MACBETH and others \citep{Saaty1977asmf,Rezaei2015bwmc,Kulakowski2014hrea,BanaECosta2016otmf}.
In these methods, the results of the comparisons constitute the decision-making
data that are subject to further processing. Let $A=\{a_{1},\ldots,a_{n}\}$
be a finite set of alternatives (available options that each expert
can choose) and $E=\{e_{1},\ldots,e_{k}\}$ be the set of experts
involved in the decision-making process. Similarly, let $C_{q}=\left\{ c_{ijq}\in\mathbb{R}_{+}\,:\,i,j=1,\ldots,n\right\} $
be a set of pairwise judgments provided by the $q$-th expert, so
that $c_{ijq}$ is the relative importance of $a_{i}$ with respect
to $a_{j}$ according to the opinion of $e_{r}$. It is convenient
to represent the set of judgments as a pairwise comparisons (PC) matrix
$C_{q}=(c_{ijq})$. For the sake of readability, however, we will
try to leave the additional index $q$ wherever it is not necessary,
i.e. when the expert's number is irrelevant. In such a case, the PC
matrix takes the form $C=(c_{ij})$. PC matrix entries can be interpreted
as ratios of individual priorities. Thus, when for some PC matrix
$C$ it holds that $c_{ij}=x$, we mean that our expert decided that
$a_{i}$ is $x$ times more important than $a_{j}$. For the same
reason, $c_{ij}=1$ means that both compared alternatives are equally
preferred. The diagonal of $C$ contains the results of comparisons
of alternatives with themselves, i.e. it is filled with $1$'s. Similarly,
in most cases we may expect that $c_{ij}=c_{ji}^{-1}$, i.e. that
the matrix is \emph{reciprocal}.
\begin{defn}
A PC matrix $C=(c_{ij})$ is said to be reciprocal if for every $c_{ij}$
holds $c_{ij}=c_{ji}^{-1}$. 
\end{defn}

The purpose of the decision-making methods is to prepare recommendations.
It usually takes the form of a numerical ranking that assigns some
real-number values to the alternatives.
\begin{defn}
Let $A$ be a set of alternatives. The numerical ranking function
for $A$ is a mapping $w:A\rightarrow\mathbb{R}_{+}$ assigning a
real and positive number to each alternative. 

The numerical ranking takes the form of a weight (priority) vector
$w$:
\[
w=\left[w(a_{1}),\ldots,w(a_{n})\right]^{T}.
\]

In the literature we may find more than a dozen methods allowing us
to determine the priority vector \citep{Kulakowski2020utahp,Mazurek2023aipc}.
The most popular ones are EVM (Eigenvalue Method) and GMM (Geometric
Mean Method) \citep{Saaty1977asmf,Crawford1985anot}. According to
the former, the ranking vector is calculated as the normalized principal
eigenvector. Thus, having the solution of equation

\[
C\widehat{w}=\lambda_{\textit{max}}\widehat{w}
\]
where $\lambda_{\textit{max}}$ is a principal eigenvalue of $C$,
the entries of the priority vector $w_{\textit{ev}}=\left[w_{\textit{ev}}(a_{1}),\ldots,w_{\textit{ev}}(a_{n})\right]^{T}$
are given as 
\[
w_{\textit{ev}}(a_{i})=\frac{\widehat{w}(a_{i})}{\sum_{j=1}^{n}\widehat{w}(a_{j})}.
\]
In the case of the latter, although the assumptions of the procedure
are similar \citep{Kulakowski2016srot}, the calculations are simpler.
In this approach, the entries of a priority vector $w_{\textit{gm}}=\left[w_{\textit{gm}}(a_{1}),\ldots,w_{\textit{gm}}(a_{n})\right]^{T}$
have the form:
\[
w_{\textit{gm}}(a_{i})=\frac{s_{i}}{\sum_{j=1}^{n}s_{j}},
\]
where 
\[
s_{i}=\left(\prod_{j=1}^{n}c_{ij}\right)^{1/n}.
\]
Thus, the individual priorities of alternatives are just geometric
means of rows of a PC matrix. Both of the aforementioned methods have
their incomplete versions \citep{Harker1987amoq,kulakowski2020otgm},
i.e. procedures that allow to calculate the priority vector even if
not all entries of $C$ are known.
\end{defn}

\subsection{Inconsistency}

Comparing alternatives pairwise is easier than comparing more alternatives
at the same time. However, if one makes comparisons independently,
it may lead to inconsistency (and it usually does).
\begin{defn}
A PCM $C$ is \textit{consistent} \citep{Kulakowski2020utahp} if
for every $i,j,k=1,\ldots,n$ holds 
\[
c_{ik}c_{kj}=c_{ij}.
\]
 It is fairly easy to prove that it is equivalent to the existence
of a positive vector $w$ such that for every $i,j=1,\ldots,n$ 
\begin{equation}
c_{ij}=\frac{w(a_{i})}{w(a_{j})}.\label{cons}
\end{equation}
\end{defn}

In real-world applications, inconsistent PCMs appear naturally. Nonetheless,
the level of inconsistency should not be too high, as it may lead
to several problems. These may include impaired sensitivity of data
\citep{Kulakowski2019tqoi}, or lack of trust in the experts' competence,
leading to the method being perceived as unreliable. Therefore plenty
of inconsistency indicators have been defined in the literature. One
of the most popular is the Consistency Index introduced by Saaty in
\citet{Saaty1977asmf}:
\begin{defn}
The Consistency Index of a $n\times n$ PC matrix $C=[c_{ij}]$ is
given by 
\[
\textit{CI}(C)=\frac{\lambda_{max}-n}{n-1},
\]
where $\lambda_{\textit{max}}$ is the principal right eigenvalue
of $C$ (i.e. the maximum one according to the absolute value). 
\end{defn}

The index $\textit{CI}(C)$ is zero if $C$ is consistent. Otherwise,
it is positive, and more specifically $0\leq\textit{CI}(C)\leq(q-1)^{2}/2q$
where $1/q\leq c_{ij}\leq q$, for $i,j=1,\ldots,n$ \citep[p. 103]{Aupetit1993osup,Kulakowski2020utahp}.

Another interesting inconsistency indicator has been proposed by Koczkodaj
\citep{Koczkodaj1993ando}. 
\begin{defn}
The Koczkodaj's inconsistency index of PC matrix $C$ is defined as
\[
K(C)=\max\left\{ K_{ijk}:1\leq i<j<k\leq n\right\} ,
\]

where 
\[
K_{ijk}=\min\left\{ \left|1-\frac{c_{ik}c_{kj}}{c_{ij}}\right|,\left|1-\frac{c_{ij}}{c_{ik}c_{kj}}\right|\right\} .
\]
\end{defn}

The difference between the two indices is that the latter is not related
to the priority deriving method, while the first one contains a reference
to the principal eigenvector of $C$. Both indices have their versions
for incomplete matrices \citep{Kulakowski2020iifi}. Very often, $K(C)$
is considered a \emph{local} inconsistency indicator, while $\textit{CI(C)}$
is called \emph{global} consistency index \citep{Kulakowski2020utahp}.

Ranking vectors can be compared in many ways. Depending on whether
the comparison is quantitative or qualitative, the appropriate metric
is used. A convenient way to compare two ordinal ranking vectors is
the Kendall Tau distance \citep{Kendall1938anmo,Kulakowski2019tqoi}. 
\begin{defn}
The Kendall Tau distance for two ordinal ranking vectors $u$ and
$v$ is defined as 
\begin{equation}
K_{\textit{rd}}(u,v)=\frac{2K_{d}(u,v)}{n(n-1)},\label{eq:norm-kendall-dist}
\end{equation}

where 
\begin{equation}
K_{d}(u,v)=\left|\left\{ (i,j)\,\text{s.t.}\,i<j\,\wedge\,\textit{sign}\left(u(a_{i})-u(a_{j})\right)\neq\textit{sign}\left(v(a_{i})-v(a_{j})\right)\right\} \right|.\label{eq:kendall-dist}
\end{equation}
\end{defn}

$K_{d}(u,v)$ counts the number of pairwise swaps that distinguish
two vectors. For example if $u=(1,2,3)$ and $v=(2,1,3)$ then $K_{d}(u,v)=1$,
as only one swap between $1$ and $2$ is needed to transform vector
$v$ to vector $u$. Due to the similarity of this idea to the so-called
bubble sort algorithm, this value is sometimes called the bubble-sort
distance.

It is easy to see that for the most distant two vectors $u$ and $v$
where $\left|u\right|=\left|v\right|=n$, the value of $K_{d}(u,v)=n(n-1)/2$
(in such a case, $u$ contains the elements of $v$ in reverse order).
Thus, the normalized Kendall distance of two ordinal vectors takes
its final form (\ref{eq:norm-kendall-dist}), where for the most distant
vectors $u,v$ the value $K_{\textit{rd}}(u,v)=1$. It is worth noting
that $K_{\textit{rd}}$ does not depend on the number of alternatives
in the ranking.

For quantitative rankings, any measure of vector distance can be used
to determine their distance. For the purpose of this article we use
the Manhattan distance; however, the Chebyshev distance \citep{Harker1987ipci}
is also common in the literature.
\begin{defn}
The Manhattan distance between two cardinal ranking vectors $u$ and
$v$ is defined as
\[
M_{d}(u,v)=\sum_{i=1}^{n}\left|u(a_{i})-v(a_{i})\right|.
\]
\end{defn}

Providing that all the entries of both $u$ and $v$ sum up to $1$,
the result $M_{d}(u,v)\leq2$.

\subsection{Group Decision Making\label{subsec:Group-Decision-Making}}

In a situation where many experts work on a recommendation and each
of them presents their own PC matrix, these data must be aggregated.
Typically, arithmetic or geometric weighted averages are used for
aggregation, although there are strong axiomatic arguments for using
geometric mean for this purpose \citep{Aczel1983pfsr}. Hence, in
our further considerations, we will focus on this very method.

We can aggregate either entire PC matrices or priority vectors resulting
from these matrices. The first approach is called AIJ (Aggregation
of Individual Judgments) while the second is referred to as AIP (Aggregation
of Individual Priorities \citep{Forman1998aija}). Let us consider
a group of experts $E=\{e^{1},\ldots,e^{k}\}$ whose task is to compare
a set $A=\{a_{1},\ldots,a_{n}\}$ of alternatives pairwise. Each of
them provides a PC matrix $C_{q}=[c_{ijq}]$ containing its personal
judgements on elements of $A$. In the AIJ approach, we begin by creating
the aggregated matrix

\[
C=\left[\begin{array}{cccc}
1 & \prod_{q=1}^{k}c_{12q}^{r_{q}} & \cdots & \prod_{q=1}^{k}c_{1nq}^{r_{q}}\\
\vdots & 1 & \vdots & \vdots\\
\vdots & \vdots & \ddots & \vdots\\
\prod_{q=1}^{k}c_{n1q}^{r_{q}} & \cdots & \cdots & 1
\end{array}\right],
\]
where $r_{1},\ldots,r_{k}\in[0,1]$ and $\sum_{q=1}^{k}r_{q}=1$.
Then, adopting $C$ as input, we calculate the final priority vector
using the method we prefer. The values $r_{1},\ldots,r_{k}$ represent
the priorities assigned to individual experts. They correspond to
the strength of the influence of individual experts' opinions on the
final result. In a situation where the opinion of each of the experts
has the same impact (which is usually the case), $r_{q}=1/k$ for
$q=1,\ldots,k$.

In the AIP approach, first, for each $C_{q}$, we calculate a priority
vector 
\[
w_{q}=\left[w_{q}(a_{1}),\ldots,w_{q}(a_{n})\right]^{T}.
\]
Then, we aggregate the vectors so that the resulting ranking is given
as 
\[
w=\left[w(a_{1}),\ldots,w(a_{n})\right]^{T},
\]
where
\begin{equation}
w(a_{i})=\frac{\prod_{q=1}^{k}w_{q}^{r_{q}}(a_{i})}{\sum_{i=1}^{n}\prod_{q=1}^{k}w_{q}^{r_{q}}(a_{i})}.\label{eq:aij-final-weight}
\end{equation}
 Similarly as before, the higher value of $r_{q}$, the stronger impact
of the $q$-th expert on the final recommendation.

\section{Inconsistency-Driven Pairwise Ranking Aggregation\label{sec:Inconsistency-Driven-Pairwise}}

\subsection{Problem statement\label{subsec:Problem-statement}}

In decision-making methods, it is common to assume the honesty and
professionalism of experts. According to the first of these assumptions,
each of the experts will try to express opinions that are as close
to the actual state as possible. In other words, they will not express
an opinion that is contrary to their own knowledge and inner conviction.
Their alleged professionalism, on the other hand, allows us to believe
that the assessment made by experts will be reliable and based on
a possibly objective comparison of various considered options. Both
of these assumptions allow us to hope that the judgments of different
yet honest and professional experts should rather coincide. Therefore,
outliers are likely to be either dishonest or unprofessional. In either
case, there is good reason to reduce the impact of such opinions on
the final result.

Describing the facts and fiction about AHP \citep[p. 22]{Forman1993fafa},
Forman notes that ``It is possible to be perfectly consistent but
consistently wrong.'' Paraphrasing this quote, one might say that
it is possible to be perfectly consistent and completely dishonest.
Nevertheless, in practice, when there are many alternatives and little
time to decide, the questions are asked in a random order and the
expert does not have the opportunity to learn about the set of alternatives
beforehand, the risk of obtaining consistent but insincere answers
does not seem very high. Hence, we can expect the insincere expert
to give answers less consistent than the average. This leads to the
formulation of a second possible heuristic that may point to dishonest
or incompetent experts. Too high an inconsistency in an expert's response
may be a good reason to reduce their impact on the final ranking values.

The above observations allow us to propose two procedures for prioritizing
experts, so that potentially dishonest experts receive a lower priority
than others. The first procedure will be based on the deviation from
the average (i.e. it will detect outliers in terms of the preferences
presented) (Sec. \ref{subsec:Distance-driven-expert-prioritiz}).
The second one will take the inconsistency in the context of a certain
average inconsistency (Sec. \ref{subsec:Inconsistency-driven-expert-prio})
into account. We are also considering a combination of both of these
heuristics (Sec. \ref{subsec:Inconsistency-driven-expert-prio-1}).

\subsubsection*{Example\label{aggr}}

One of the common variants of manipulation in the social choice theory
is control \citep{Brandt2016hocs}. It is usually carried out by the
election organizer. Paradigmatic examples of control include adding
or deleting voters. A similar effect can be seen with the pairwise
comparison method. Hence, by adding or removing experts, one can try
to affect the ranking results\footnote{In contrast to the rank reversal problem \citep{Dyer1990rota}, in
this case we are dealing with deliberately crafted experts' answers
.}. Let us consider the example where the opinion of six experts $e_{1},\ldots,e_{6}$
was taken into account in order to develop recommendations on the
four alternatives under consideration. The experts' opinions were
in the form of $4\times4$ matrices:

\noindent 
\[
C_{1}=\left(\begin{array}{cccc}
1. & 1.322 & 1.926 & 2.0784\\
0.7566 & 1. & 1.63 & 2.6057\\
0.5192 & 0.6136 & 1. & 1.011\\
0.4811 & 0.3838 & 0.9888 & 1.
\end{array}\right),
\]

\[
C_{2}=\left(\begin{array}{cccc}
1. & 1.335 & 2.408 & 3.2864\\
0.749 & 1. & 2.0619 & 1.619\\
0.4153 & 0.485 & 1. & 1.599\\
0.3043 & 0.6178 & 0.6255 & 1.
\end{array}\right),
\]

\[
C_{3}=\left(\begin{array}{cccc}
1. & 1.034 & 2.0437 & 2.6168\\
0.9668 & 1. & 1.956 & 1.987\\
0.4893 & 0.5113 & 1. & 1.32\\
0.3821 & 0.5034 & 0.7574 & 1.
\end{array}\right),
\]

\[
C_{4}=\left(\begin{array}{cccc}
1. & 1.3 & 2.1679 & 2.0552\\
0.7691 & 1. & 1.603 & 2.5181\\
0.4613 & 0.624 & 1. & 1.005\\
0.4866 & 0.3971 & 0.9949 & 1.
\end{array}\right),
\]
\[
C_{5}=\left(\begin{array}{cccc}
1. & 1.052 & 2.1005 & 2.5668\\
0.9501 & 1. & 2.025 & 1.646\\
0.4761 & 0.4938 & 1. & 1.564\\
0.3896 & 0.6074 & 0.6394 & 1.
\end{array}\right),
\]
\[
C_{6}=\left(\begin{array}{cccc}
1. & 1.153 & 2.2591 & 2.4132\\
0.8669 & 1. & 1.832 & 1.629\\
0.4426 & 0.5459 & 1. & 1.284\\
0.4144 & 0.6137 & 0.7785 & 1.
\end{array}\right).
\]

\noindent The normalized weight vectors obtained by GMM are as follows:
\[
w_{1}=\left[0.355802,0.314085,0.176758,0.153356\right]^{T},
\]

\noindent 
\[
w_{2}=\left[0.409827,0.285829,0.171231,0.133114\right]^{T},
\]
\[
w_{3}=\left[0.356504,0.323638,0.176235,0.143623\right]^{T},
\]
\[
w_{4}=\left[0.362966,0.31053,0.171584,0.154919\right]^{T},
\]
\[
w_{5}=\left[0.360622,0.311708,0.181946,0.145724\right]^{T},
\]
\[
w_{6}=\left[0.371263,0.297355,0.174992,0.156391\right]^{T},
\]

\noindent which means that the six experts indicate $a_{1}$ as the
best alternative. Their aggregated ranking is 
\[
w_{1-6}=\left[0.369045,0.306942,0.175421,0.147627\right]^{T}.
\]

\noindent However, a dishonest organizer (process facilitator) added
two more experts $e_{7}$ and $e_{8}$ who lobby for $a_{2}$. As
they know that $a_{1}$ is its main competitor, they proposed opinions
(matrices $C_{7}$ and $C_{8}$) i.e.: 

\noindent 
\[
C_{7}=\left(\begin{array}{cccc}
1. & 0.1752 & 0.5622 & 0.3212\\
5.7078 & 1. & 2.584 & 1.604\\
1.779 & 0.387 & 1. & 1.785\\
3.1135 & 0.6236 & 0.5602 & 1.
\end{array}\right),
\]
\[
C_{8}=\left(\begin{array}{cccc}
1. & 0.4551 & 0.2515 & 0.5273\\
2.1971 & 1. & 2.0847 & 2.2497\\
3.9758 & 0.4797 & 1. & 1.47\\
1.896 & 0.4445 & 0.6803 & 1.
\end{array}\right),
\]

\noindent resulting in a ranking downgrading $a_{1}$ and boosting
$a_{2}$:

\noindent 
\[
w_{7}=\left[0.0897127,0.469102,0.223955,0.21723\right]^{T},
\]

\noindent 
\[
w_{8}=\left[0.0897127,0.469102,0.223955,0.21723\right]^{T}.
\]
Then, applying the standard aggregation process \citep{Forman1998aija,Groselj2015cosa},
the final ranking is as follows: 
\[
w_{1-8}=\left[0.266227,0.334807,0.192645,0.160465\right]{}^{T}.
\]
This determines the order of alternatives: $a_{2},a_{1},a_{3},a_{4}$
which is in line with the expectations of experts $e_{7}$ and $e_{8}$.

\subsection{Preferential distance-driven expert prioritization\label{subsec:Distance-driven-expert-prioritiz}}

Let $\widehat{w}_{i}=\left[\widehat{w}_{i}(a_{1}),\ldots,\widehat{w}_{i}(a_{k})\right]^{T}$
be the ranking vector calculated using either EVM or GMM (Sec. \ref{subsec:Pairwise-comparisons}),
based on the matrix $C_{i}$ provided by the expert $e_{i}$ for $i=1,\ldots,k$.
Similarly, let $\widehat{w}=\left[\widehat{w}(a_{1}),\ldots,\widehat{w}(a_{n})\right]^{T}$
be a ranking vector calculated using AIP (Sec. \ref{subsec:Group-Decision-Making}),
based on $\widehat{w}_{1},\ldots,\widehat{w}_{k}$. According to the
adopted heuristics, the more the opinion of the $i$-th expert differs
from that of the entire team, the greater the risk of manipulation.
Thus, let 
\begin{equation}
\textit{d}_{i}=\left|\widehat{w}-\widehat{w}_{i}\right|\label{eq:indiv-distances}
\end{equation}
be a quantitative distance\footnote{For example, Manhattan distance, Euclidean distance or Chebyshev distance.}
between vectors $\widehat{w}$ and $\widehat{w}_{i}$. Then, we need
to map the individual distances $d_{i}$ (\ref{eq:indiv-distances})
to priority values over a certain numerical scale. For this purpose,
let us denote the minimum and maximum of the values $D=\{d_{1},\ldots,d_{k}\}$
as $d_{\textit{min}}$ and $d_{\textit{max}}$ correspondingly. As
$d_{\textit{min}}$ corresponds to the most preferred expert and $d_{\textit{min}}$
corresponds to the least preferred expert, we assign the value $h\in\mathbb{R}_{+}$
to $d_{\textit{min}}$ and $l\in\mathbb{R}_{+}$ to $d_{\textit{max}}$,
where of course $h>l$. The ratio $h/l$ should correspond to the
ratio between the reliability of the expert corresponding to $d_{\textit{min}}$
to the reliability of the expert corresponding to $d_{\textit{max}}$.
Values $l$ and $h$ form a scale to which all the distances $d_{1},\ldots,d_{k}$
will be transformed.

Let $f:\mathbb{R}_{+}\rightarrow\mathbb{R}$ be a mapping transforming
distances $d_{i}$ to priorities, which after normalization can be
used in the weighted AIP procedure. Function $f$ should pass through
two points $X=(d_{\textit{min}},h)$ and $Y=(d_{\textit{max}},l)$,
so that the highest priority value $h$ is assigned to the expert
whose opinion was closest to the mean, and the lowest priority value
$l$ is assigned to the expert whose opinion was the farthest from
the mean.

The mapping $f$ lets us use a linear function passing through two
points $X=(x_{1},x_{2})$ and $Y=(y_{1},y_{2})$ in the form 
\begin{equation}
f_{XY}(x)=\frac{x_{2}-y_{2}}{x_{1}-y_{1}}x+\left(x_{2}-\frac{x_{2}-y_{2}}{x_{1}-y_{1}}x_{1}\right).
\end{equation}
This allows us to calculate the values\footnote{As long as points $X$ and $Y$ are well-defined, we will write $f(x)$
instead of $f_{XY}(x)$.} $f(d_{1}),\ldots,f(d_{k})$ which determines the priorities of experts
$e_{1},\ldots,e_{k}$. In order to satisfy the form of the weighted
geometric mean, one would need to rescale the priority values so that
they sum up to one. Hence, the final experts' priorities take the
form:

\begin{equation}
r_{i}=\frac{f(d_{i})}{\sum_{i=1}^{k}f(d_{i})},\,\,\,i=1,\ldots,k.\label{eq:prior_resc}
\end{equation}
After computing $r_{1},\ldots,r_{k}$, we calculate the final priority
vector $w=\left[w(a_{1}),\right.$ $\ldots$ $\left.,w(a_{n})\right]^{T}$
using the weighted version of AIP (\ref{eq:aij-final-weight}).

\begin{figure}
\begin{centering}
\includegraphics[width=0.9\textwidth]{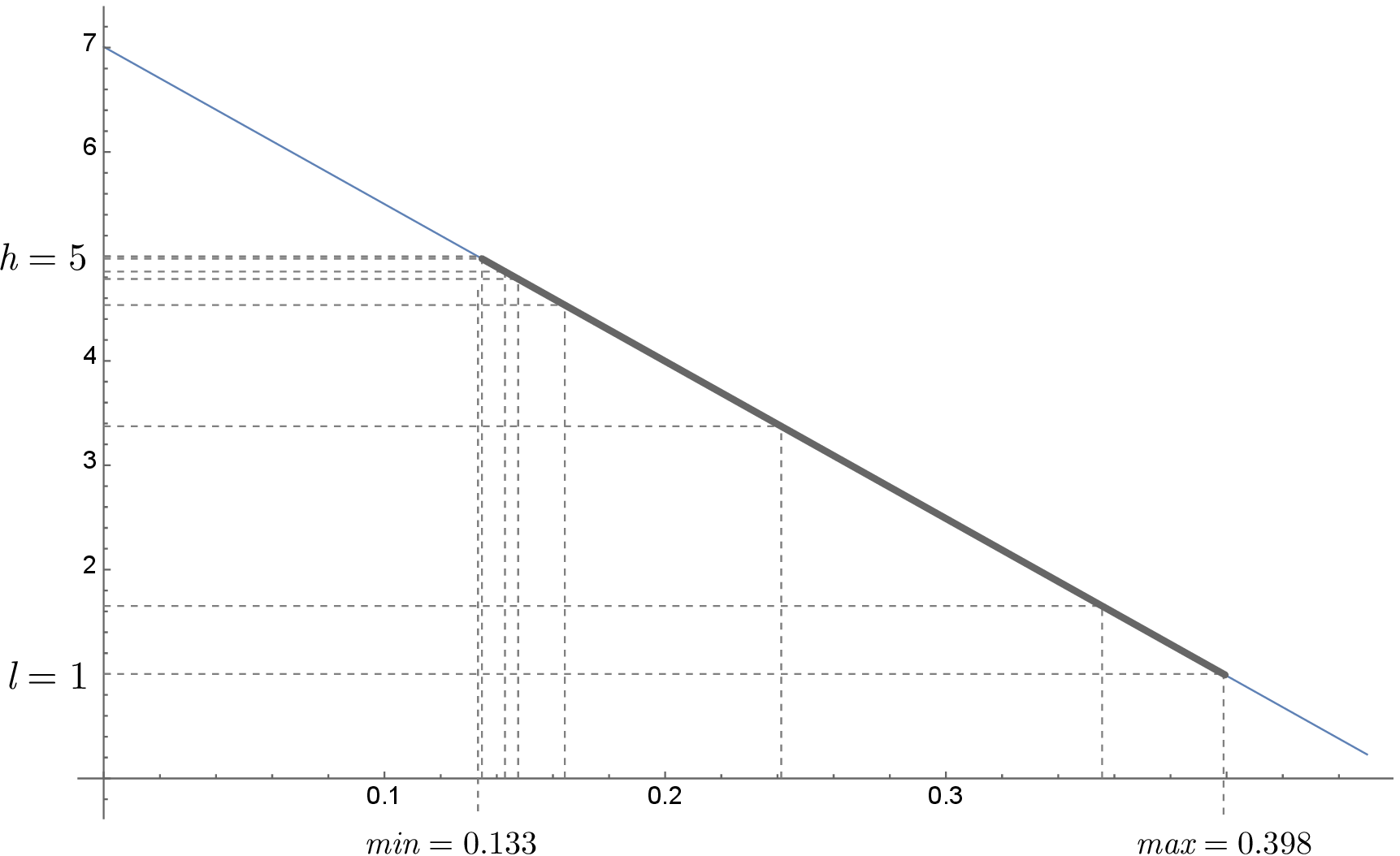}
\par\end{centering}
\caption{Linear function mapping distances to priorities passing points $(min,h)$
and $(max,l)$. }
\label{fig:linear-mappin-example}
\end{figure}

For the purpose of the aforementioned example (see end of Section
\ref{subsec:Problem-statement}), let us assume that the expert with
the lowest value $d_{\textit{min}}$ is strongly more credible than
the expert with the highest value $d_{\textit{max}}$. Following the
fundamental scale the ratio $h/l=5$, so we may assign $h=5$ and
$l=1$. As a result, we get the linear mapping function $f(x)=-15.0607x+7.0075$,
determining the experts' weights (Fig. \ref{fig:linear-mappin-example}).

The weight values are therefore: 
\[
\left[f(d_{1}),\ldots,f(d_{8})\right]^{T}=\left[5.,3.37,4.97,4.78,4.85,4.53,1.,1.65\right]^{T}.
\]
Rescaling produces the values that can be used as the input to the
AIP procedure: 
\[
\left[r_{1},\ldots,r_{8}\right]^{T}=\left[0.165,0.11,0.16,0.15,0.16,0.15,0.0331,0.054\right]^{T}.
\]
After re-aggregating the results, we get a priority vector:
\[
w_{1-8}=\left[0.327,0.317,0.182,0.152\right]^{T}.
\]
As we can see, the alternative $a_{1}$ with the priority $w_{1-8}(a_{1})=0.327$,
which is preferred by the majority of voters, returned to the winning
position.

\subsection{Inconsistency-driven expert prioritization\label{subsec:Inconsistency-driven-expert-prio}}

Instead of using the distance between individual judgment vectors
and its mean value, we may use the distance between the inconsistencies
of individual experts and the average inconsistency of their judgments.
Thus, let 
\[
d_{i}=\left(I(C_{i})-\frac{1}{k}\sum_{j=1}^{k}I(C_{j})\right),
\]

where $I$ is some selected inconsistency index \citep{Brunelli2018aaoi,Kulakowski2020iifi}.
Contrary to the previous case, in which the distance of the individual
ranking from the average value was important, here we have to measure
whether the inconsistency of a given expert is below or above the
average. If it is above average, it may mean either an attempt (perhaps
a naive one) to manipulate the deficiencies of the expert himself
(lack of experience, lack of firmness, distraction, time pressure,
etc.). However, if the individual ranking is below average, i.e.,
the expert's consistency is higher than the average, this may, to
some extent, be in favor of the expert. On the other hand, a certain
degree of inconsistency is considered desirable \citep[p. 265]{Saaty2008rmai}
or \citep[p. 172]{Saaty1987tahp}. In other words, both too high and
too small a degree of inconsistency may indicate strategic decision-making,
although it is quite difficult to ``punish'' too much consistency.

These two observations lead to the conclusion that the mapping of
distances to priorities should differ depending on the sign of $d_{i}$.
Providing that at least two matrices $C_{i}$ and $C_{j}$ such that
$I(C_{i})\neq I(C_{j})$ exist, there must exist also two matrices
such that $I(C_{p})<1/k\sum_{j=1}^{k}I(C_{j})<I(C_{\textit{q}})$.
Let the inconsistency values for the most consistent $e_{\textit{min}}$
and inconsistent $e_{\textit{max}}$ expert be: $I_{\textit{min}}=\min_{j=1,\ldots,k}I(C_{j})$
and $I_{\textit{max}}=\max_{j=1,\ldots,k}I(C_{j})$. Additionally,
let $I_{\textit{mid}}$ be the value of inconsistency of the expert
whose opinions' consistency is closest to the average, i.e. $I_{\textit{mid}}=I(C_{i})$
such that $\left|I(C_{i})-1/k\sum_{j=1}^{k}I(C_{j})\right|$ is minimal.

In the next step, we should set the high, middle and low weights,
i.e. $h,m$ and $l$ corresponding to the values of $I_{\textit{min}},I_{\textit{mid}}$
and $I_{\textit{max}}$. Thus, we perform three comparisons of experts'
credibility, which result in the following matrix: 
\begin{equation}
\textit{C}_{\textit{ex}}=\left(\begin{array}{ccc}
1 & \textit{c}_{\textit{min},\textit{mid}} & \textit{c}_{\textit{min},\textit{max}}\\
1/\textit{c}_{\textit{min},\textit{mid}} & 1 & \textit{c}_{\textit{mid},\textit{max}}\\
1/\textit{c}_{\textit{min},\textit{max}} & 1/\textit{c}_{\textit{mid},\textit{max}} & 1
\end{array}\right).\label{eq:tree-point-credibility-matrix}
\end{equation}
It is worth noting that in this heuristic, the smaller the inconsistency,
the better the results, assuming that $\textit{c}_{\textit{min},\textit{mid}}$,
$\textit{c}_{\textit{mid},\textit{max}}$ and $\textit{c}_{\textit{min},\textit{max}}$
cannot be greater than $1$.

Calculating a ranking based on (\ref{eq:tree-point-credibility-matrix})
brings us the desired values of $h,m$ and $l$. These allow us to
formulate three points through which the function mapping the distance
$d_{i}$'s to the weight values of individual experts should pass.
These points are: $A=(d_{\textit{min}},h)$, $B=(d_{\textit{mid}},m)$
and $C=(d_{\textit{max}},l)$. 

As mapping $f:\mathbb{R}_{+}\rightarrow\mathbb{R}$ transforming distances
$I_{i}$ to priorities let us use a piecewise linear function including
two segments: $A-B$ and $B-C$. Thus, 
\[
f_{\textit{XYZ}}(x)=\begin{cases}
f_{\textit{XY}}(x) & x<0\\
f_{\textit{YZ}}(x) & x\geq0
\end{cases}.
\]
Similarly as before, $f_{\textit{ABC}}=f$ allows us to calculate
the values $f(d_{1}),\ldots,f(d_{k})$ which, after appropriate rescaling
(\ref{eq:prior_resc}), form the experts' priorities $r_{1},\ldots,r_{k}$.
Finally, the ranking is calculated, taking into account the priorities
of individual experts.

\begin{figure}
\begin{centering}
\includegraphics[width=0.9\columnwidth]{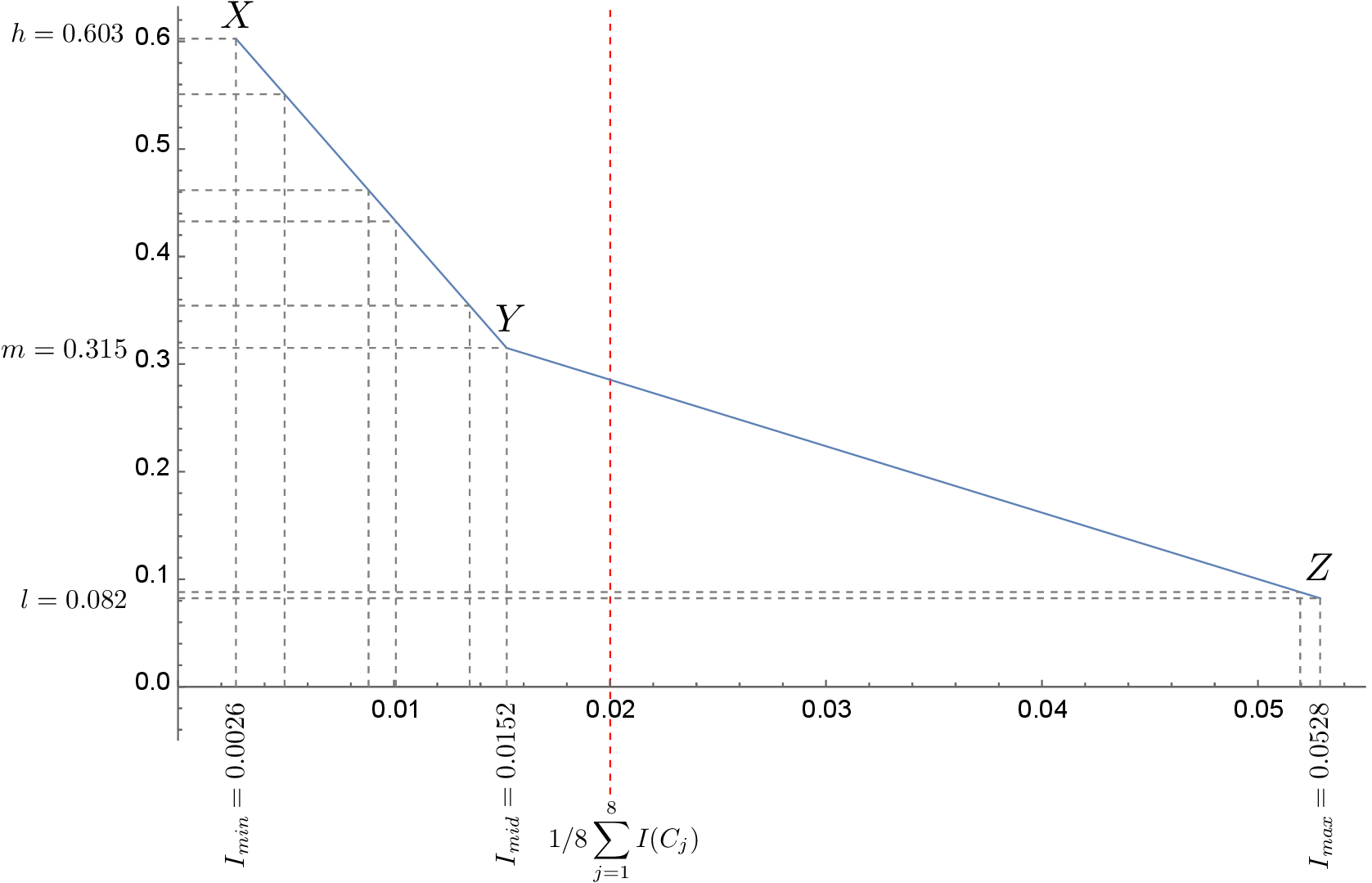}
\par\end{centering}
\caption{Piecewise linear function mapping distances to priorities passing
points $A,B$ and $C$.}
\label{fig:linear-mappin-example-2}
\end{figure}

In the case of Example \ref{aggr} experts $e_{1},\ldots,e_{8}$ achieved
the following consistency values:
\[
w_{\textit{cexp}}=\{0.0101,0.0152,0.0026,0.0088,0.013,0.0049,0.0528,0.0519\}
\]
It is easy to see that $I_{\textit{min}}=0.0026$ (expert $e_{3}$)
and $I_{\textit{max}}=0.0528$ (expert $e_{7}$). The average inconsistency
is $0.02$, thus the nearest inconsistency result was achieved by
the expert $e_{2}$ with $I_{\textit{mid}}=0.0152$. In the next step,
we need to compare the credibility of $e_{3},e_{2}$ and $e_{7}$.
Let 
\[
C_{\textit{ex}}=\left(\begin{array}{ccc}
1 & 2 & 7\\
\frac{1}{2} & 1 & 4\\
\frac{1}{7} & \frac{1}{4} & 1
\end{array}\right),
\]
which yields the following credibility values: $0.603,0.315$ and
$0.082$. This allows us to construct the mapping $f_{XYZ}$ (Fig.
\ref{fig:linear-mappin-example-2}) where $A=(0.0026,0.603)$, $B=(0.02,0.315)$
and $C=(0.082,0.082)$.

According to the function $f_{\textit{XYZ}}=f$, experts $e_{1},\ldots,e_{8}$
obtain the following weights: 
\[
\left[f(d_{1}),\ldots,f(d_{8})\right]^{T}=[0.432,0.315,0.602,0.462,0.354,0.551,0.082,0.088]^{T}.
\]
After rescaling so that all weights sum up to $1$, we obtain the
values that can be used as the input to the AIP procedure:

\[
\left[r_{1},\ldots,r_{8}\right]^{T}=\left[0.149,0.109,0.208,0.159,0.122,0.19,\ 0.028,0.03\right]^{T}.
\]
Then, re-aggregation of experts' results leads to the final priority
vector:
\[
w_{1-8}=\left[0.339,0.314,0.1793,0.151\right]^{T}.
\]

The alternative $a_{1}$ with the priority $w_{1-8}(a_{1})=0.329$
has the highest rank, whilst $a_{2}$ correctly takes the second place.
Thanks to the introduction of priorities, it was again possible to
mitigate the manipulation.

\subsection{Mixed expert prioritization\label{subsec:Inconsistency-driven-expert-prio-1}}

It is possible to use both of the above heuristics simultaneously.
So let $r_{i}^{1}$ be the weight of the $i$-th expert calculated
on the basis of the heuristic distance from the mean judgment (Sec.
\ref{subsec:Distance-driven-expert-prioritiz}), while $r_{i}^{2}$
is the weight of the expert calculated from the difference in inconsistencies
(Sec. \ref{subsec:Inconsistency-driven-expert-prio}). Thus, the mixed
expert weight is the linear combination of both: 
\[
r_{i}=\beta r_{i}^{1}+(1-\beta)r_{i}^{2},\,\,\text{for}\,\,i=1,\ldots,k,
\]
where $0\leq\beta\leq1$ is the coefficient determining the impact
of both heuristics. Since $\sum_{i=1}^{k}r_{i}^{1}=\sum_{i=1}^{k}r_{i}^{2}=1$,
then also $\sum_{i=1}^{k}r_{i}=1$. Thus, the obtained weights $r_{1},\ldots,r_{k}$
fit the definition of the weighted geometric mean.

In case of Example \ref{aggr} and assuming that both heuristics contribute
equally to the weights of the experts, i.e. $\beta=0.5$, we get

\[
\left[r_{1},\ldots,r_{8}\right]^{T}=\left[0.157,0.11,0.186,0.159,0.141,0.17,\ 0.03,0.042\right]^{T},
\]
which results in the following priority vector: 
\[
w_{1-8}=\left[0.333,0.316,0.18,0.151\right]^{T}.
\]

\subsection{The degree of expert's credibility }

In each of the two heuristics described above (Section \ref{subsec:Distance-driven-expert-prioritiz}
and \ref{subsec:Inconsistency-driven-expert-prio}), two or three
key experts are first selected for credibility assessment. Then, based
on this result, a mapping is proposed to prioritize all experts. The
credibility evaluation must be made in accordance with the adopted
heuristics, i.e. experts less consistent in their judgments or more
distant from the average than the competitor have to get the lower
score. This may result in a certain inconvenience for those assessing
the credibility of experts. They may focus on their attitude towards
individual experts, and not on the quality of their expertise. As
a result, a person who is liked and popular in the society may get
a better assessment than an expert who is reliable but not sociable.

The way to avoid this trap is to provide a procedure that allows us
to prioritize key experts without explicitly comparing them. Then,
we can assume a priori that the ratio of the best to the worst expert
(the first heuristic, Section \ref{subsec:Distance-driven-expert-prioritiz})
is e.g. $5:1$, or ratios of the best, average and worst experts (the
second heuristic, Section \ref{subsec:Inconsistency-driven-expert-prio})
is e.g. $9:4:1$.

It is also possible to determine these relations in a procedural/functional
way. For instance, in the case of our example and the second heuristic,
we may assume that the expert priority should be linearly correlated
with inconsistency. Thus, as $d_{\textit{min}}=0.0026$, $d_{\textit{mid}}=0.0152$
and $d_{\textit{max}}=0.0528$ (Section \ref{subsec:Inconsistency-driven-expert-prio})
the priority may take the values: $h=\alpha\cdot d_{\textit{max}}/d_{\textit{min}}$,
$m=\alpha\cdot d_{\textit{mid}}/d_{\textit{min}}$ and $l=1$, where
$\alpha\geq1$ is a gain factor.

\section{Montecarlo experiments\label{sec:Montecarlo-experiments}}

\subsection{Data preparation}

For the purpose of both experiments (Sections \ref{subsec:Defense-against-manipulation}
and \ref{subsec:Vulnerability-to-original}), we prepared $4,000$
samples of $20$-matrix sets corresponding to different decision scenarios.
For this purpose, we first drew $34$ priority vectors for $5$ alternatives,
$33$ vectors for $6$ alternatives, and another $33$ vectors for
$7$ alternatives. Then, as every priority vector corresponds to exactly
one consistent PC matrix \citep{Kulakowski2020utahp}, we created
$100$ consistent PC matrices of sizes $5\times5,6\times6$ and $7\times7$.
Thus, if 
\[
w=[w(a_{1}),\ldots,w(a_{5})]^{T}
\]
is a priority vector corresponding to certain five alternatives, then
the consistent matrix corresponding to $w$ is: 
\[
C_{w}=\left(\begin{array}{ccccc}
1 & \frac{w(a_{1})}{w(a_{2})} & \frac{w(a_{1})}{w(a_{3})} & \frac{w(a_{1})}{w(a_{4})} & \frac{w(a_{1})}{w(a_{5})}\\
\frac{w(a_{2})}{w(a_{1})} & 1 & \frac{w(a_{2})}{w(a_{3})} & \frac{a_{2})}{w(a_{4})} & \frac{w(a_{2})}{a_{5})}\\
\frac{w(a_{3})}{w(a_{1})} & \frac{w(a_{3})}{w(a_{2})} & 1 & \frac{w(a_{3})}{w(a_{4})} & \frac{w(a_{3})}{w(a_{5})}\\
\frac{w(a_{4})}{w(a_{1})} & \frac{w(a_{4})}{w(a_{2})} & \frac{w(a_{4})}{w(a_{3})} & 1 & \frac{w(a_{4})}{w(a_{5})}\\
\frac{w(a_{5})}{w(a_{1})} & \frac{w(a_{5})}{w(a_{2})} & \frac{w(a_{5})}{w(a_{3})} & \frac{w(a_{5})}{w(a_{4})} & 1
\end{array}\right).
\]

In the next step, we disturbed the elements of these matrices by multiplying
them by a random factor $\epsilon\in[1/\alpha,\alpha],$ where $\alpha=1.1,1.2,\ldots,5$.
Thus, the disturbed version of $C_{w}$ takes the form: 

\[
\widetilde{C}_{w,\alpha}=\left(\begin{array}{ccccc}
1 & \epsilon_{12}\frac{w(a_{1})}{w(a_{2})} & \epsilon_{13}\frac{w(a_{1})}{w(a_{3})} & \epsilon_{14}\frac{w(a_{1})}{w(a_{4})} & \epsilon_{15}\frac{w(a_{1})}{w(a_{5})}\\
\epsilon_{21}\frac{w(a_{2})}{w(a_{1})} & 1 & \epsilon_{23}\frac{w(a_{2})}{w(a_{3})} & \epsilon_{24}\frac{a_{2})}{w(a_{4})} & \epsilon_{25}\frac{w(a_{2})}{a_{5})}\\
\epsilon_{31}\frac{w(a_{3})}{w(a_{1})} & \epsilon_{32}\frac{w(a_{3})}{w(a_{2})} & 1 & \epsilon_{34}\frac{w(a_{3})}{w(a_{4})} & \epsilon_{35}\frac{w(a_{3})}{w(a_{5})}\\
\epsilon_{41}\frac{w(a_{4})}{w(a_{1})} & \epsilon_{42}\frac{w(a_{4})}{w(a_{2})} & \epsilon_{43}\frac{w(a_{4})}{w(a_{3})} & 1 & \epsilon_{45}\frac{w(a_{4})}{w(a_{5})}\\
\epsilon_{51}\frac{w(a_{5})}{w(a_{1})} & \epsilon_{52}\frac{w(a_{5})}{w(a_{2})} & \epsilon_{53}\frac{w(a_{5})}{w(a_{3})} & \epsilon_{54}\frac{w(a_{5})}{w(a_{4})} & 1
\end{array}\right),
\]

where $\epsilon_{ij}\in[1/\alpha,\alpha]$ and $\epsilon_{ij}=1/\epsilon_{ji}$
for $i,j=1,\ldots,5$. For every consistent PC matrix $C_{w_{x}}$
(where $x=1,\ldots,100$) and $\alpha_{k}\in\{1.1,\ldots,5\}$ we
randomly selected $20$ matrices $\widetilde{C}_{w,\alpha_{k}}^{(q)}$
where $q=1,\ldots,20$. Thus, every set $S_{w_{x}\alpha_{k}}=\left\{ \widetilde{C}_{w_{x},\alpha_{k}}^{(1)},\ldots,\widetilde{C}_{w_{x},\alpha_{k}}^{(q)}\right\} $
corresponds to a single group decision scenario, in which $q=20$
experts make decisions as to the priorities of $5,6$ or $7$ alternatives
convergent (to some extent) with the vector $w_{x}$. The input to
each experiment were $4,000$ such $S_{w_{x}\alpha_{k}}$ corresponding
to different sets of alternatives ($100$ vectors $w_{x}$) and different
average inconsistency of experts ($40$ different values of $\alpha_{k}$).

For every $S_{w_{x}\alpha_{k}}$ we determined the average level of
inconsistency as the arithmetic mean of its components, i.e. 
\[
I(S_{w_{x}\alpha_{k}})=\frac{I\left(\widetilde{C}_{w_{x},\alpha_{k}}^{(1)}\right)+\ldots+I\left(\widetilde{C}_{w_{x},\alpha_{k}}^{(q)}\right)}{\left|S_{w_{x}\alpha_{k}}\right|},
\]
where $I$ denotes the inconsistency indicator (in the experiments
we used Saaty's consistency index $\textit{CI}$) and $q=\left|S_{w_{x}\alpha_{k}}\right|$
is the number of experts involved in the decision-making process (in
the experiments, we adopted the number 20). As a general rule, an
increase in $\alpha_{k}$ causes an increase in $I(S_{w_{x}\alpha_{k}})$.

For the purpose of the experiments, we used GMM (Section \ref{subsec:Pairwise-comparisons})
to calculate the priorities. In such case, it is easy to show that
AIP and AIJ (Section \ref{subsec:Group-Decision-Making}) lead to
the same results, therefore we do not need to consider both aggregation
methods separately.

\subsection{Defense against manipulation\label{subsec:Defense-against-manipulation}}

\subsubsection{Model of manipulation\label{subsec:Model-of-manipulation}}

In the first experiment, we will assume that a certain number of experts
is bribed to submit manipulated matrices. For the purposes of the
experiment, we assume that the grafter's goal is to make the original
second alternative the winner of the ranking. For this purpose, he
bribes several experts who are most in favor of the original winner
(bribing experts who do not support the current winner seems to be
a less effective strategy). In exchange for a bribe, the experts undertake
to indicate that the comparison of the alternative being promoted
by the grafter with any other is $9$ (the largest value of the fundamental
scale), and the comparison of the current leader with any other alternative
is $1/9$.

Let us see how this somewhat simple group decision-making manipulation
scheme works on the simple example. In order to evaluate the five
alternatives, four different experts would prepare four pairwise comparison
matrices:
\[
C_{1}=\left(\begin{array}{ccccc}
1. & 0.246 & 3.731 & 2.75 & 0.65\\
4.069 & 1. & 10.83 & 9.54 & 1.25\\
0.268 & 0.092 & 1. & 0.39 & 0.311\\
0.363 & 0.105 & 2.566 & 1. & 0.729\\
1.54 & 0.799 & 3.219 & 1.37 & 1.
\end{array}\right),
\]

\[
C_{2}=\left(\begin{array}{ccccc}
1. & 0.347 & 1.03 & 0.668 & 0.629\\
2.881 & 1. & 7.614 & 5.696 & 0.835\\
0.969 & 0.131 & 1. & 1.98 & 0.335\\
1.5 & 0.175 & 0.506 & 1. & 0.734\\
1.59 & 1.2 & 2.982 & 1.36 & 1.
\end{array}\right),
\]

\[
C_{3}=\left(\begin{array}{ccccc}
1. & 0.798 & 2.497 & 0.61 & 0.359\\
1.25 & 1. & 4.799 & 1.57 & 1.29\\
0.4 & 0.208 & 1. & 0.374 & 0.544\\
1.64 & 0.637 & 2.672 & 1. & 1.01\\
2.787 & 0.774 & 1.84 & 0.99 & 1.
\end{array}\right),
\]

\[
C_{4}=\left(\begin{array}{ccccc}
1. & 0.83 & 1.28 & 1.91 & 0.535\\
1.2 & 1. & 8.778 & 7.832 & 3.302\\
0.781 & 0.114 & 1. & 1.09 & 0.244\\
0.524 & 0.128 & 0.921 & 1. & 0.228\\
1.87 & 0.303 & 4.097 & 4.376 & 1.
\end{array}\right).
\]
which would result in the following four priority vectors: 
\[
w_{1}=[0.16,0.507,0.045,0.085,0.203]^{T},
\]
\[
w_{2}=[0.115,0.425,0.102,0.105,0.252]^{T},
\]

\[
w_{3}=[0.154,0.301,0.081,0.224,0.24]^{T},
\]

\[
w_{4}=[0.154,0.467,0.072,0.065,0.242]^{T}.
\]
After aggregation via the AIJ method the ranking vector would be:
\[
w=[0.145,0.417,0.072,0.107,0.233]^{T},
\]
i.e. the winner is the second alternative with the score $w(a_{2})=0.417$.
Therefore, in order to push through the candidature of the vice-leader
of the current ranking, i.e. alternative $a_{5}$ with the score $w(a_{5})=0.233$,
the grafter bribes $a_{2}$'s the strongest supporter, i.e. the expert
no. 1. Hence, in fact, the first expert submits a manipulated matrix:
\[
\widetilde{C}_{1}=\left(\begin{array}{ccccc}
1 & 9 & 3.731 & 2.75 & 1/9\\
1/9 & 1 & 1/9 & 1/9 & 1/9\\
0.268 & 9 & 1 & 0.39 & 1/9\\
0.363 & 9 & 2.566 & 1 & 1/9\\
9 & 9 & 9 & 9 & 1
\end{array}\right).
\]
After aggregation of $\widetilde{C}_{1},C_{2},C_{3}$ and $C_{4}$
it turns out that the final priorities have the following values:
\[
\widetilde{w}=[0.148,0.183,0.08,0.113,0.31]^{T},
\]
which means that the manipulation was successful. The new winner was
alternative $a_{5}$ with the score $w(a_{5})=0.31$, which without
manipulation would have taken the second place. If bribing one expert
was not enough, the grafter would try to bribe next strongest supporter
of $a_{2}$, etc.

In the above procedure, we assume that the grafter knows who the winner's
strongest supporter is (and bribery of whom could potentially be most
disadvantageous to the winner and beneficial to the preferred alternative).
In practice, the grafter usually does not have such knowledge and
must rely on his intuition and knowledge of expert preferences. So
one can hope that, in practice, a real grafter will work less efficiently
than the one from our experiment.

\subsubsection{Experiment results}

The input data for the experiment included $4,000$ sets $S_{w_{x}\alpha_{k}}$
composed of twenty PC matrices, each corresponding to the given initial
priority vector $w_{x}$ and the range of disturbance factor $\alpha_{k}$.
For each set $S_{w_{x}\alpha_{k}}$ first we calculate the aggregated
priorities\footnote{For the sake of simplicity and clarity, we assume that the best alternative
in the non-manipulated ranking is indexed as $a_{1}$, the second
best alternative is $a_{2}$ and so on.}

\begin{equation}
\bar{w}_{x}=\left[\bar{w}_{x}(a_{1}),\bar{w}_{x}(a_{2}),\ldots,\bar{w}_{x}(a_{n})\right]^{T},\label{eq:ranking-with-standard-aip}
\end{equation}
and based on it we carry out a simulated manipulation attack in accordance
with the method described in Section \ref{subsec:Model-of-manipulation}.
We also determine the average inconsistency of the experts' responses
$I\left(S_{w_{x}\alpha_{k}}\right)$. As expected, along the increase
of the $\alpha_{k}$ coefficient, the average inconsistency also gets
higher.

In most cases, it is enough to ``bribe'' one to three experts, i.e.
manipulate between one and three matrices from the entire $S_{w_{x}\alpha_{k}}$
set. In each of the analyzed cases, the attack method is effective.
This means that in each considered case it is possible to ``improve''
the experts' answers to achieve the intended goal, i.e. to promote
the second best alternative in the original ranking, as the new leader.
Let us denote the manipulated set of expert answers as $\widetilde{S}_{w_{x}\alpha_{k}}$,
and the manipulated priority vector for $\widetilde{S}_{w_{x}\alpha_{k}}$
aggregated using the AIP method as: 
\[
\widetilde{w}_{x}=\left[\widetilde{w}_{x}(a_{2}),\widetilde{w}_{x}(a_{i_{1}}),\ldots,\widetilde{w}_{x}(a_{i_{n}})\right]^{T},
\]
where $i_{1},i_{2},\ldots,i_{n}$ is some permutation of indices from
the set $\{1,3,4,\ldots,n\}$. Of course, according to the assumption
of the manipulation (Section \ref{subsec:Model-of-manipulation}),
even though $w_{x}(a_{1})>w_{x}(a_{2})$, in the manipulated ranking
$\widetilde{w}(a_{1})<\widetilde{w}(a_{2})$. Then, we estimated the
priority vector with the help of APDD (Aggregation of Preferential
Distance-Driven expert prioritization, Section \ref{subsec:Distance-driven-expert-prioritiz}),
AID (Aggregation of Inconsistency-Driven expert prioritization, Section
\ref{subsec:Inconsistency-driven-expert-prio}) and MX, the mixed
method based on a linear combination of priorities of APDD and AID
(Section \ref{subsec:Inconsistency-driven-expert-prio-1}). As a result,
for each $\widetilde{S}_{w_{x}\alpha_{k}}$ we obtained the following
vectors: 
\[
\widehat{w}_{x}^{\textit{APDD}}=\left[\widehat{w}_{x}^{\textit{APDD}}(a_{p_{1}}),\ldots,\widehat{w}_{x}^{\textit{APDD}}(a_{p_{n}})\right]^{T},
\]

\[
\widehat{w}_{x}^{\textit{AID}}=\left[\widehat{w}_{x}^{\textit{AID}}(a_{q_{1}}),\ldots,\widehat{w}_{x}^{\textit{AID}}(a_{q_{n}})\right]^{T},
\]

\[
\widehat{w}_{x}^{\textit{MX}}=\left[\widehat{w}_{x}^{\textit{MX}}(a_{r_{1}}),\ldots,\widehat{w}_{x}^{\textit{MX}}(a_{r_{n}})\right]^{T},
\]
where $(p_{1},\ldots,p_{n})$, $(q_{1},\ldots,q_{n})$ and $(r_{1},\ldots,r_{n})$
are some permutations of the elements from $\{1,2,\ldots,n\}$. We
consider $\widehat{w}_{x}^{\textit{APDD}}$ as a WR (winner restoration
case) if the order of the first two alternatives has been restored,
i.e. $p_{1}=1$ and $p_{2}=2$, and as a RR (ranking restoration case),
if the order of the alternatives is the same as before manipulation
i.e. $p_{1}=1,p_{2}=2,\ldots,p_{n}=n$. The cases of $\widehat{w}_{x}^{\textit{APDD}}$
when $p_{1}\neq1$ or $p_{2}\neq2$ are considered as a ``failure''.
We similarly denoted results in the case of $\widehat{w}_{x}^{\textit{AID}}$
and $\widehat{w}_{x}^{\textit{MX}}$.

\begin{figure}
\begin{centering}
\includegraphics[width=0.9\textwidth]{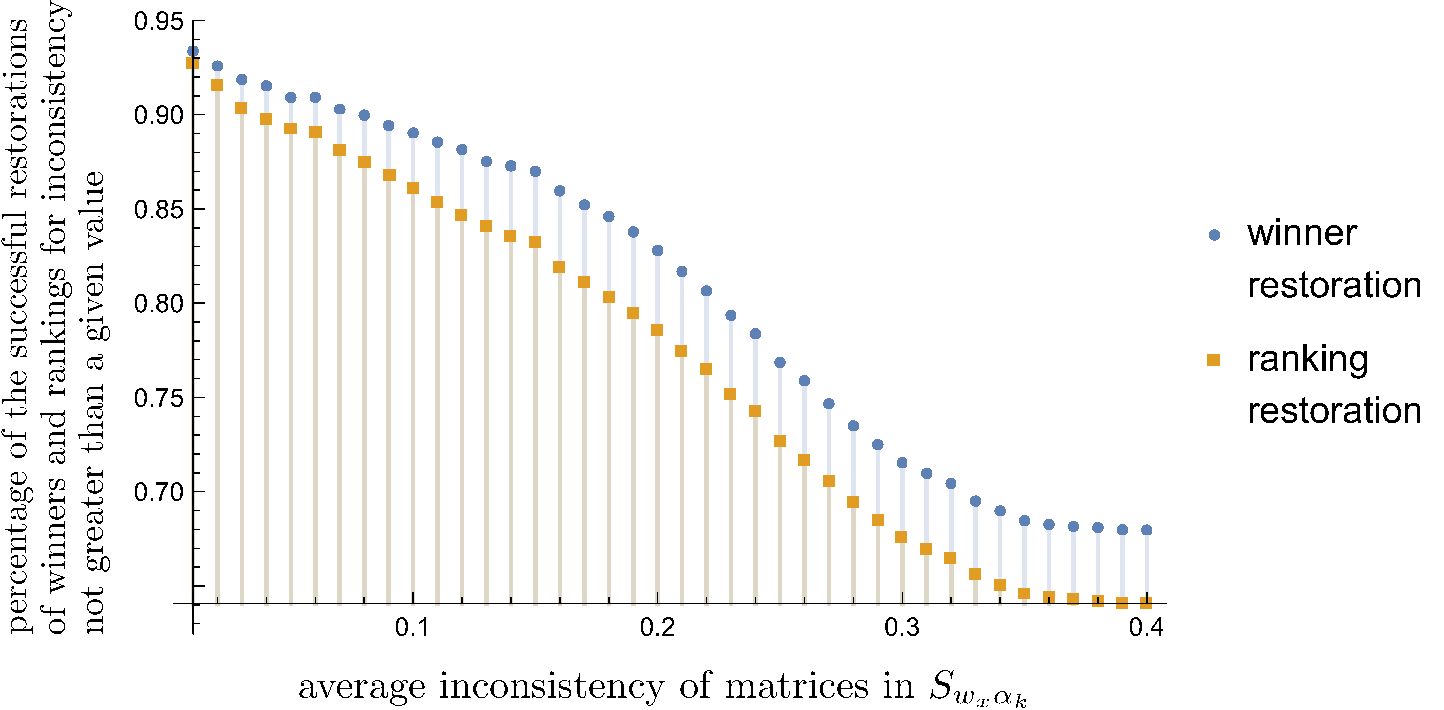}
\par\end{centering}
\caption{The ratio of number of cases where APDD managed to restore the winner
(circles) or restore the entire ranking (squares) to the number of
all considered cases.}

\label{fig:fig_3_apdd_success_ratio}
\end{figure}

In Figure \ref{fig:fig_3_apdd_success_ratio} we can see how the ratios
of WR and RR to the number of considered sets $S_{w_{x}\alpha_{k}}$
changes with the increase in the average inconsistency of $S_{w_{x}\alpha_{k}}$
in case of the APDD method. In particular, we can observe that for
the average consistency $\textit{CI}\leq0.1$, there are $89\%$ of
cases (value $0.89$ on the plot) in which APDD was able to restore
the correct winner. Similarly, there are $86\%$ of cases where APDD
reconstructed a complete ranking.

\begin{figure}
\begin{centering}
\includegraphics[width=0.9\textwidth]{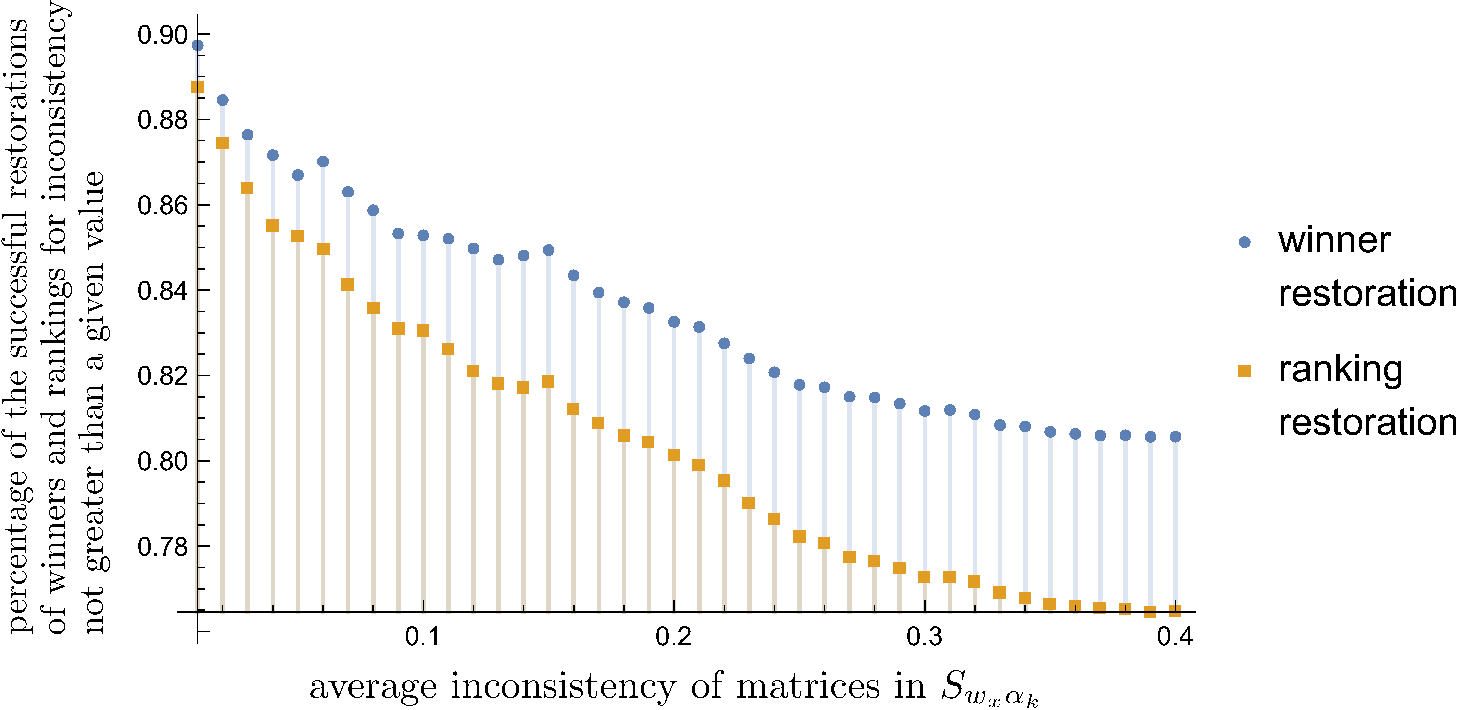}
\par\end{centering}
\caption{The ratio of number of cases where AID managed to restore the winner
(circles) or restore the entire ranking (squares) to the number of
all considered cases.}

\label{fig:fig_4_aid_success_ratio}
\end{figure}

\begin{figure}
\begin{centering}
\includegraphics[width=0.9\textwidth]{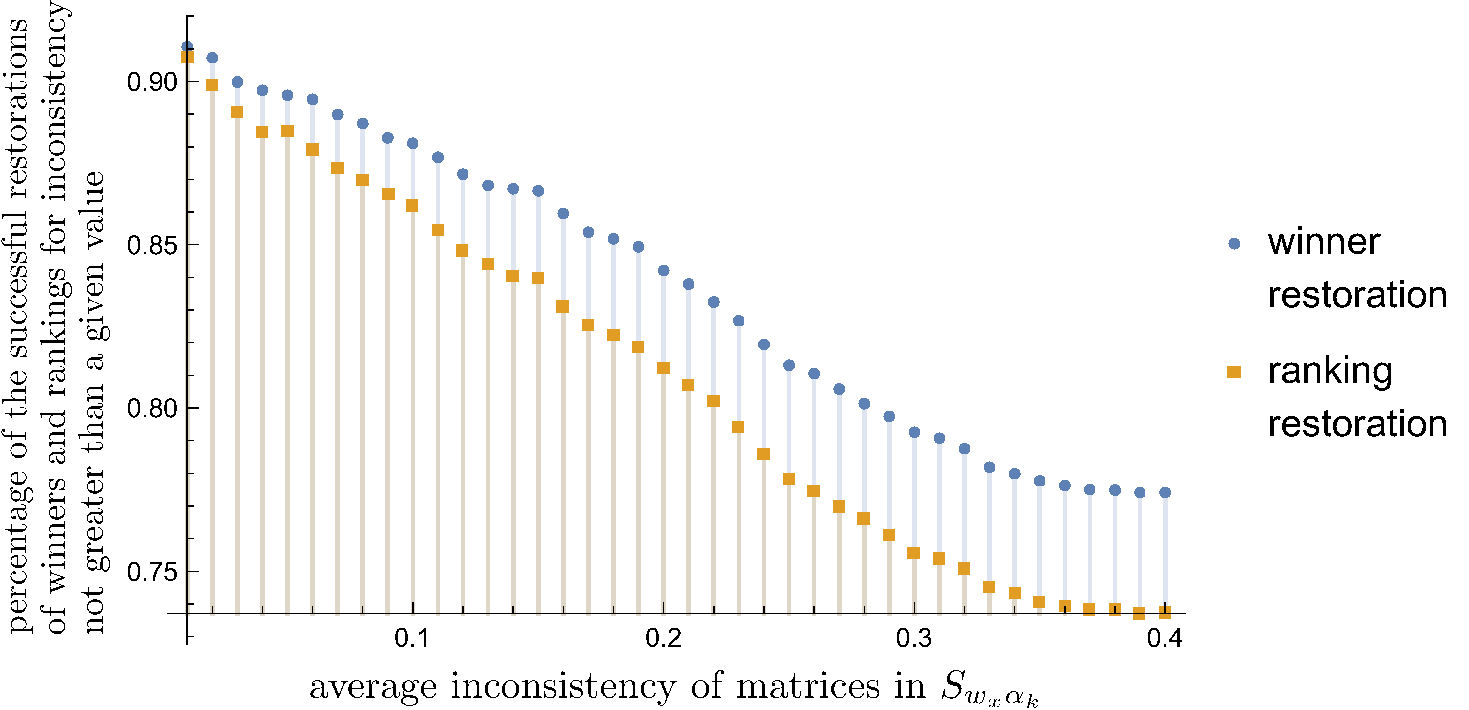}
\par\end{centering}
\caption{The ratio of number of cases where MX managed to restore the winner
(circles) or restore the entire ranking (squares) to the number of
all considered cases.}

\label{fig:fig_5_mx_success_ratio}
\end{figure}

The AID and MX methods results are presented in Figures \ref{fig:fig_4_aid_success_ratio}
and \ref{fig:fig_5_mx_success_ratio}, respectively.

For $\textit{CI}\leq0.1$, AID restored the winner in $85\%$ and
the whole ranking in $83\%$. The combined MX method reconstructed
the winer in $88\%$ and the complete ranking in $86\%$. In all cases,
the number of decision models in which the winner was restored is
slightly larger than the number of those where the entire ranking
was restored.

\begin{figure}
\begin{centering}
\includegraphics[width=0.7\textwidth]{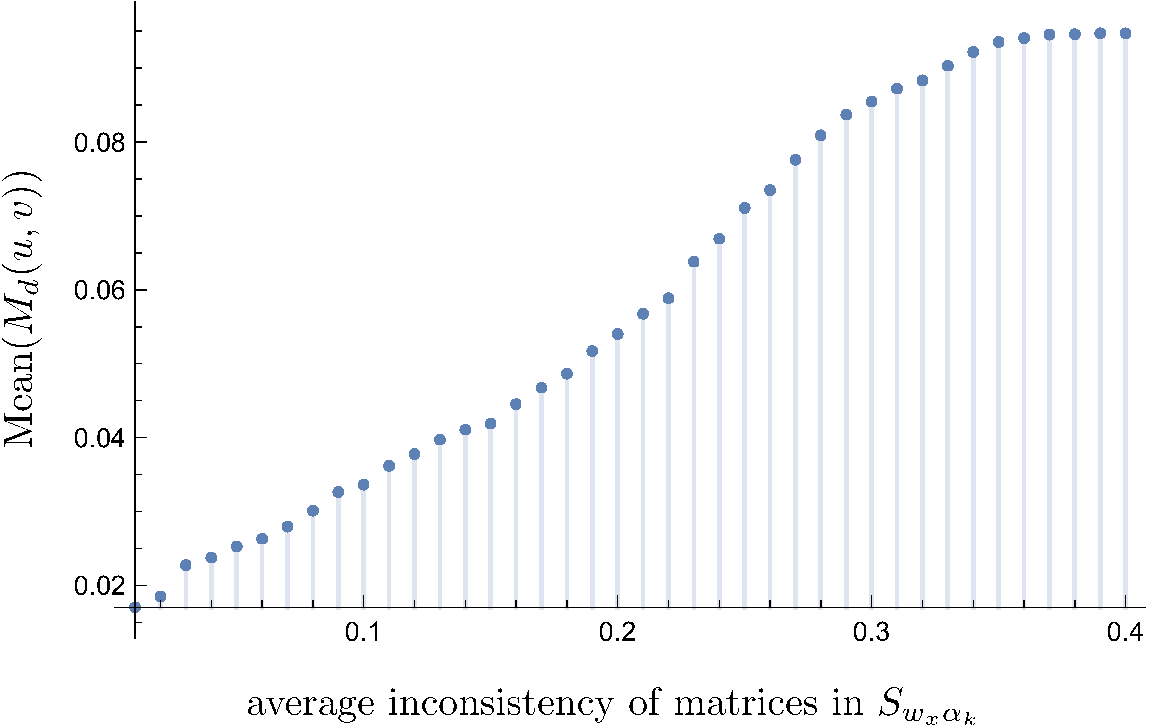}
\par\end{centering}
\caption{The average Manhattan distance $M_{d}$ between the un-manipulated
ranking and its counterpart restored by the APDD method after manipulation.}

\label{fig:fig_6_apdd_distance}
\end{figure}

\begin{figure}
\begin{centering}
\includegraphics[width=0.7\textwidth]{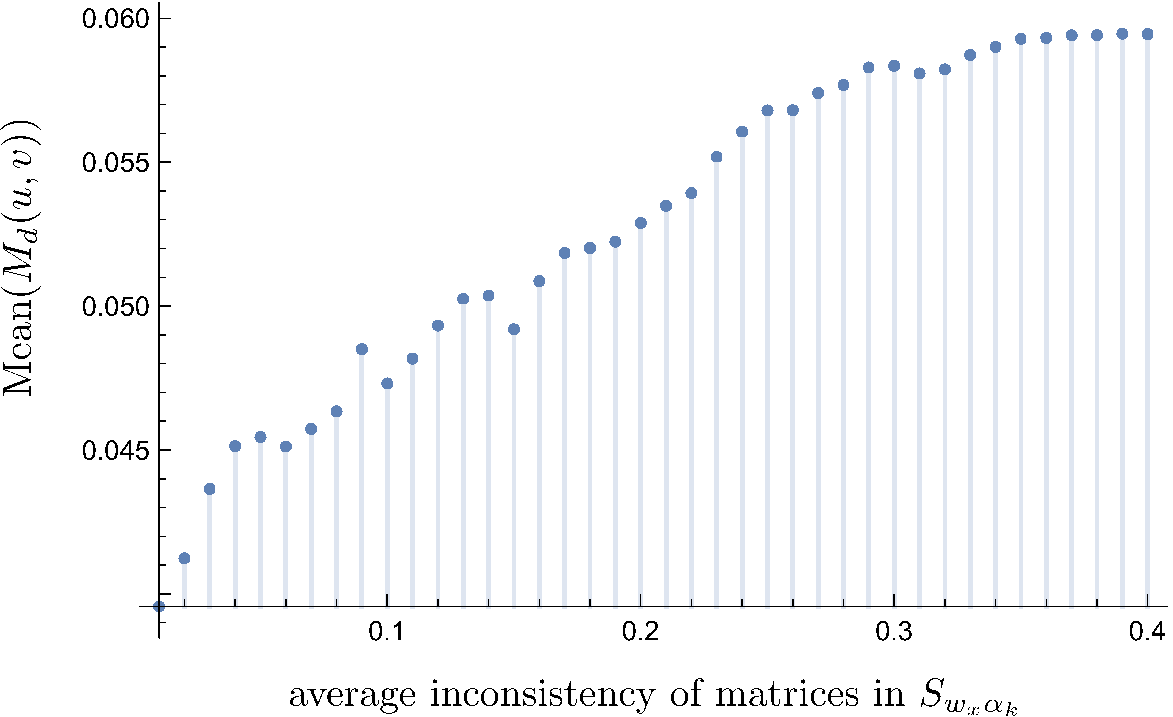}
\par\end{centering}
\caption{The average Manhattan distance $M_{d}$ between the un-manipulated
ranking and its counterpart restored by the AID method after manipulation.}

\label{fig:fig_7_aid_distance}
\end{figure}

\begin{figure}
\begin{centering}
\includegraphics[width=0.7\textwidth]{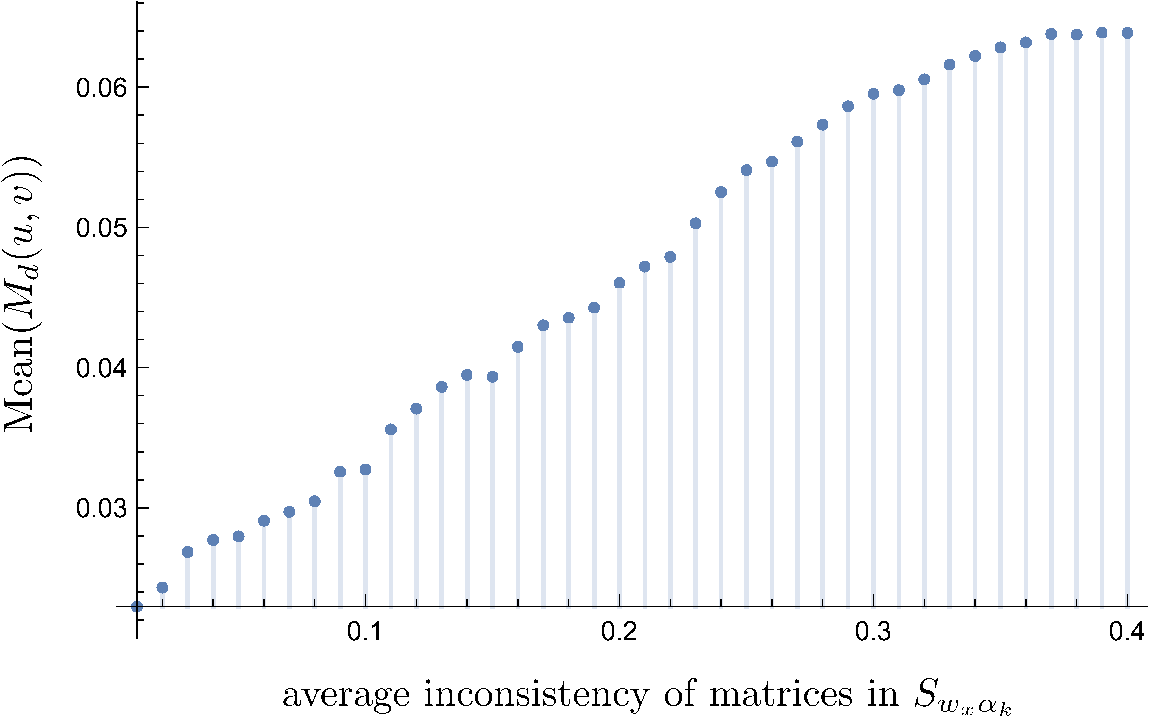}
\par\end{centering}
\caption{The average Manhattan distance $M_{d}$ between the un-manipulated
ranking and its restored by the MX method after manipulation counterpart.}

\label{fig:fig_8_mx_distance}
\end{figure}

Despite the good efficiency in restoring the order of alternatives,
the APDD, AID and MX methods are not able to ensure that the resulting
ranking will be exactly the same as before the manipulation. In this
case, the quality of the obtained result also depends on the average
inconsistency of $S_{w_{x}\alpha_{k}}$. In Figures \ref{fig:fig_6_apdd_distance},
\ref{fig:fig_7_aid_distance} and \ref{fig:fig_8_mx_distance} we
see how the restored results differ from non-manipulated rankings
expressed in the form of the average values of the Manhattan distances\footnote{$M_{d}(u,v)=1/n\sum_{i=1}^{n}\left|u(a_{i})-v(a_{i})\right|$}
depending on $M_{d}(\bar{w}_{x},\widehat{w}_{x}^{*})$, where $*$
denotes the method used, e.g. APDD, AID or MX.

It can be seen that with a relatively small average inconsistency
of experts (let's say\footnote{In \citep{Saaty1977asmf} the assessment of inconsistency was made
dependent on the average inconsistency of the random matrix. Therefore,
to decide on the acceptability of the level of inconsistency, it is
more convenient to use \emph{CR} (consistency ratio) \citep{Saaty1977asmf}.
Since the assessment of inconsistency is not the aim of the work,
we abandoned the use of \emph{CR} in favor of \emph{CI} (inconsistency
index).}, around $0.1$), the difference between the original and the reconstructed
ranking is also reasonably small. In the case of APDD it is: $M_{d}(u,v)=0.0336$,
AID: $M_{d}(u,v)=0.047$ and MX $M_{d}(u,v)=0.0327$. Therefore, in
a large number of cases, such a result can be considered acceptable.

\subsection{Vulnerability to original ranking perturbation\label{subsec:Vulnerability-to-original}}

In the second experiment, we assumed that all experts acted honestly.
Hence, methods of aggregating expert opinions are defined to minimize
the effects of manipulation one can perceive as disturbances. Therefore,
our goal this time is to check to what extent the proposed methods
can ``disturb'' the actual ranking if the manipulation did not occur.
For this purpose, we took the same dataset as in the previous experiment,
but this time we did not manipulate individual data sets $S_{w_{x}\alpha_{k}}$
to modify the original result. Similarly as before, $\bar{w}_{x}$
denotes the aggregated ranking (\ref{eq:ranking-with-standard-aip})
calculated using the standard AIP procedure (see Sec. \ref{subsec:Group-Decision-Making}).
The results aggregated using the modified APDD, AID and MX aggregation
procedures will be denoted as $\breve{w}_{x}$ with the appropriate
superscripted acronym. However, this time both ranking vectors $\bar{w}_{x}$
and $\breve{w}_{x}$ are calculated based on the same data set $S_{w_{x}\alpha_{k}}$.
Therefore, the distance between them can be understood as an indicator
of the disturbance of the original ranking by unnecessary use of the
APDD, AID and MX aggregation methods.

\begin{figure}[h]
\begin{centering}
\includegraphics[width=0.7\textwidth]{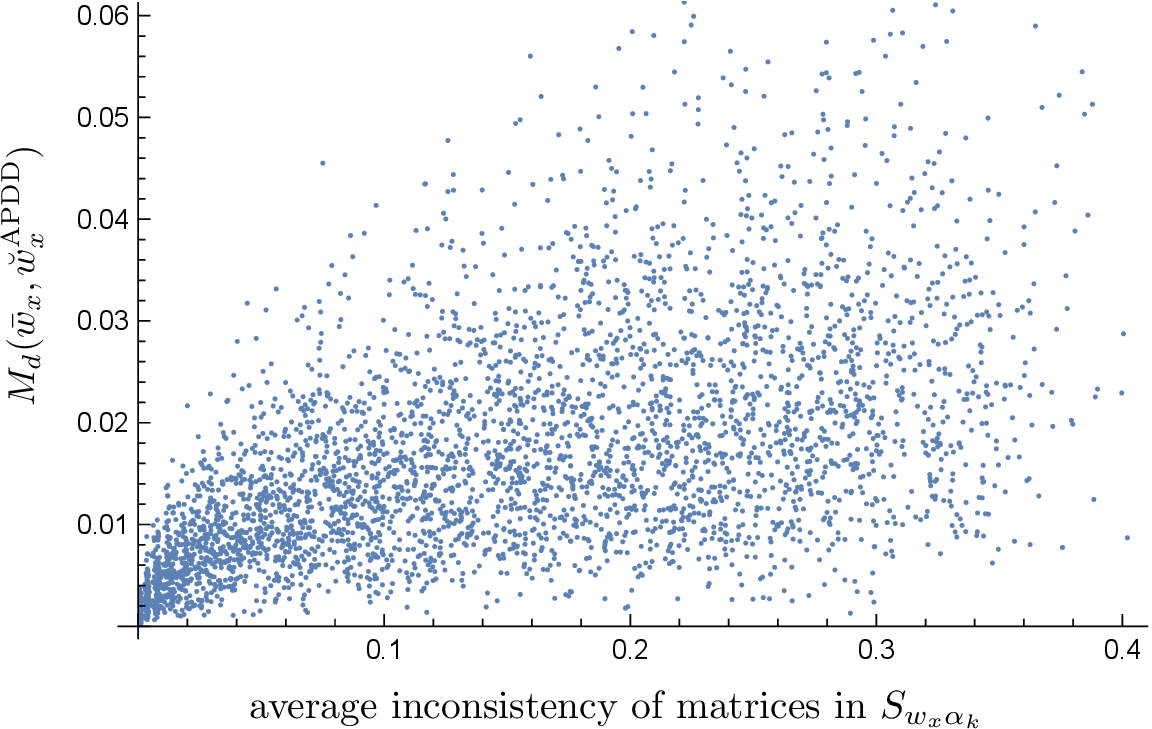}
\par\end{centering}
\caption{Distances between non-manipulated rankings calculated in a standard
way $\bar{w}_{x}$ and with the help of the APDD method.}

\label{fig:fig_9_apdd_manh_dist_vs_inconsist}
\end{figure}

\begin{figure}[h]
\begin{centering}
\includegraphics[width=0.7\textwidth]{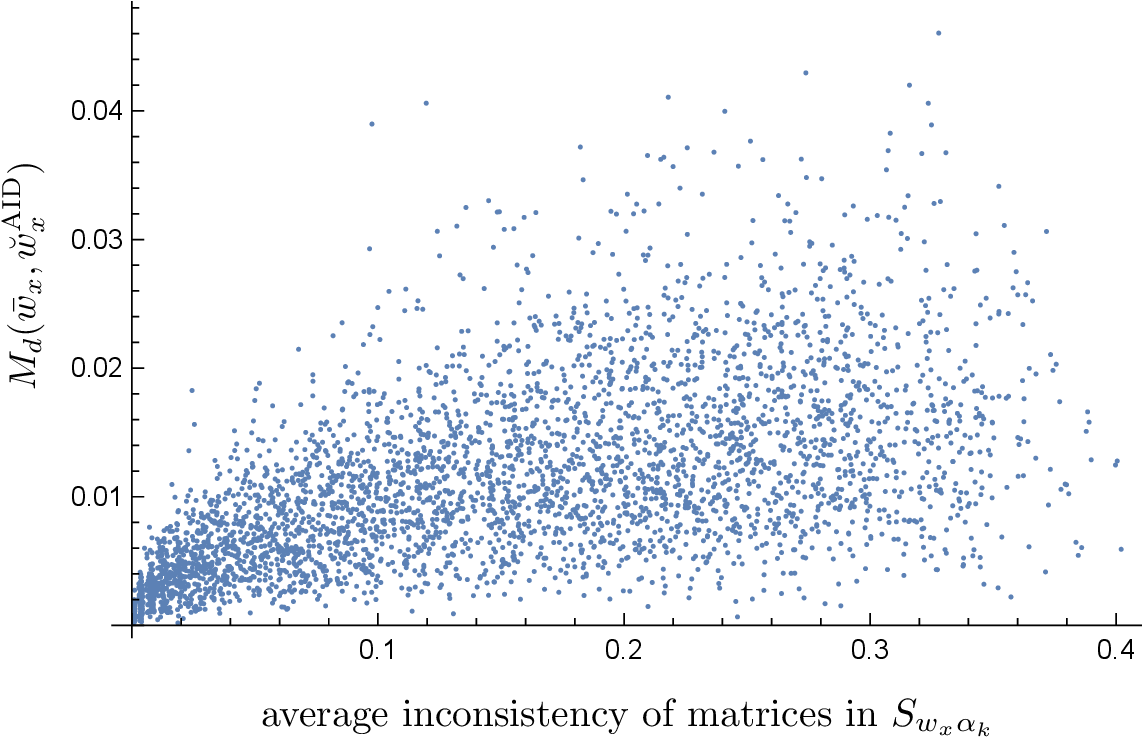}
\par\end{centering}
\caption{Distances between non-manipulated rankings calculated in a standard
way $\bar{w}_{x}$ and with the help of the AID method.}

\label{fig:fig_10_aid_manh_dist_vs_inconsist}
\end{figure}

\begin{figure}[h]
\begin{centering}
\includegraphics[width=0.7\textwidth]{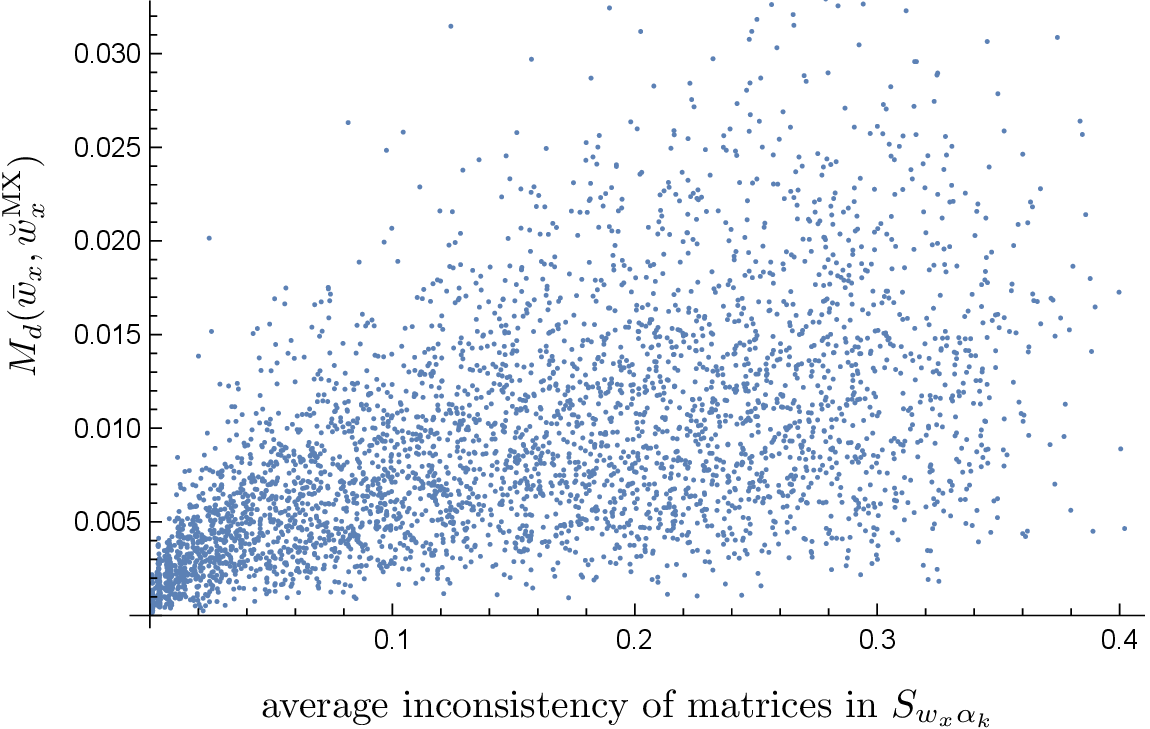}
\par\end{centering}
\caption{Distances between non-manipulated rankings calculated in a standard
way $\bar{w}_{x}$ and with the help of the MX method.}

\label{fig:fig_11_mx_manh_dist_vs_inconsist}
\end{figure}

As expected, (Fig. \ref{fig:fig_9_apdd_manh_dist_vs_inconsist}, \ref{fig:fig_10_aid_manh_dist_vs_inconsist}
and \ref{fig:fig_11_mx_manh_dist_vs_inconsist}) the size of the ranking
disturbances depends on the degree of inconsistency. 

Basically, the greater the inconsistency, the greater the differences
between the two vectors $\bar{w}_{x}$ and $\breve{w}_{x}$. Interestingly,
the best results are achieved by the mixed method (Fig. \ref{fig:fig_11_mx_manh_dist_vs_inconsist}),
which is a combination of both other strategies (Fig. \ref{fig:fig_8_mx_distance},
\ref{fig:fig_9_apdd_manh_dist_vs_inconsist}). The advantage of the
MX method can be observed not just visually. Indeed, the average distances
$M_{d}\left(\bar{w}_{x},\breve{w}_{x}\right)$ between rankings aggregated
using standard and modified procedures are as follows:
\begin{equation}
\frac{1}{4000}\sum_{\bar{w}_{x}}M_{d}\left(\bar{w}_{x},\breve{w}_{x}^{\text{APDD}}\right)=0.017\label{eq:apdd-quantitative-diff-2nd-mc-ex}
\end{equation}

\begin{equation}
\frac{1}{4000}\sum_{\bar{w}_{x}}M_{d}\left(\bar{w}_{x},\breve{w}_{x}^{\text{AID}}\right)=0.011\label{eq:aid-quantitative-diff-2nd-mc-ex}
\end{equation}

\begin{equation}
\frac{1}{4000}\sum_{\bar{w}_{x}}M_{d}\left(\bar{w}_{x},\breve{w}_{x}^{\text{MX}}\right)=0.009\label{eq:mx-quantitative-diff-2nd-mc-ex}
\end{equation}
From the above, it is easy to see that, on average, the results of
the MX method (value $0.009$) are the least distant from the unmodified
aggregation method. 

\begin{figure}[H]
\begin{centering}
\includegraphics[width=0.7\textwidth]{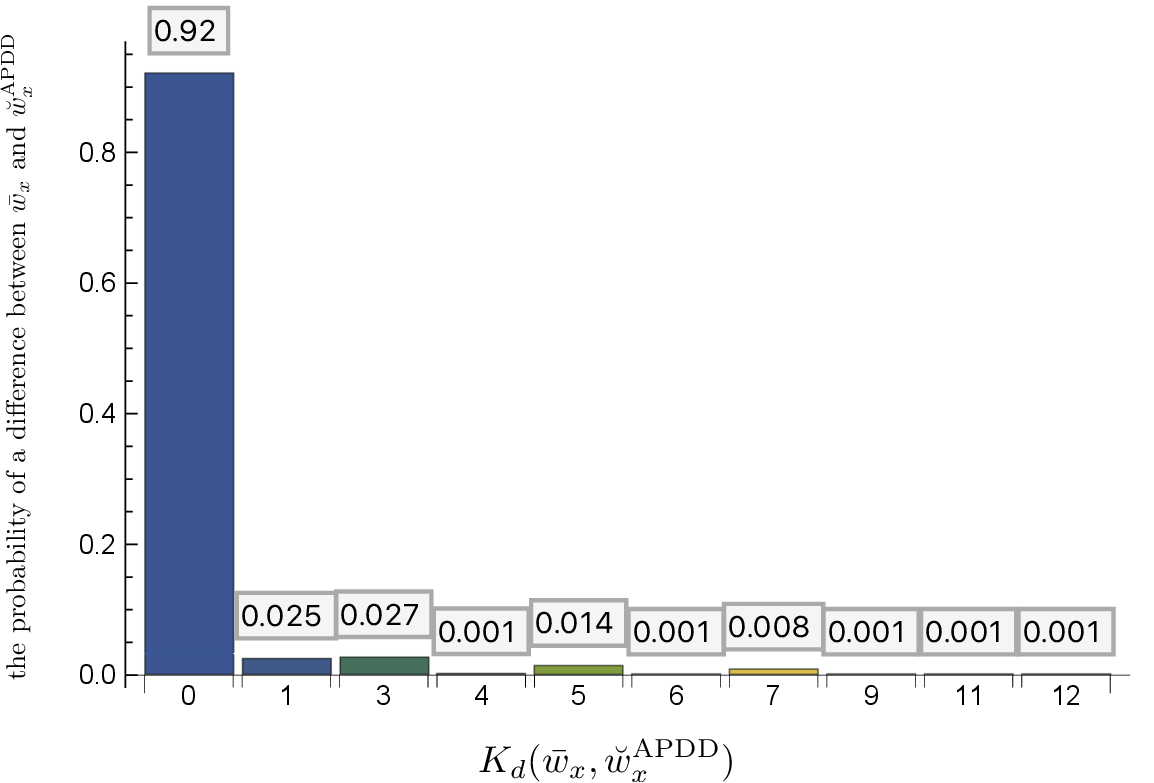}
\par\end{centering}
\caption{Estimated probability that for $I(S_{w_{x}\alpha_{k}})\protect\leq0.1$
the Kendall Tau distance $K_{d}(\bar{w}_{x},\breve{w}_{x}^{\text{APDD}})$
will be: $0,1,\ldots,12$.}

\label{fig:fig_12_apdd_kendall_prob}
\end{figure}

The disruption of the ranking may be not only quantitative, but also
qualitative. This means that the modified method may propose a ranking
that will differ from the ``original'' with regard to the order
of the alternatives. As the standard method for determining the ordinal
difference between rankings is the Kendall Tau distance (\ref{eq:kendall-dist}),
we calculated\footnote{In the experiment, the compared rankings did not have ties. Hence,
the expression (\ref{eq:kendall-dist}) could be used to calculate
the Kendall Tau distance $K_{d}$. } $K_{d}(\bar{w}_{x},\breve{w}_{x})$ for every $S_{w_{x}\alpha_{k}}$
for which $I(S_{w_{x}\alpha_{k}})\leq0.1$. The obtained values (vertical
axis) can be interpreted as the expected probability that with the
assumed not too high inconsistency of the group of experts ($I(S_{w_{x}\alpha_{k}})\leq0.1$),
both rankings will differ by a given number of transpositions (horizontal
axis). For example, in the case of the APDD method, we can see (Fig.
\ref{fig:fig_12_apdd_kendall_prob}) that if the average inconsistency
in the group of experts is not too high, $I(S_{w_{x}\alpha_{k}})\leq0.1$
then there is a $92\%$ chance that both rankings remain unchanged.
Going forward, there is a $2.5\%$ chance that they will differ in
one transposition, etc.

\begin{figure}[H]
\begin{centering}
\includegraphics[width=0.7\textwidth]{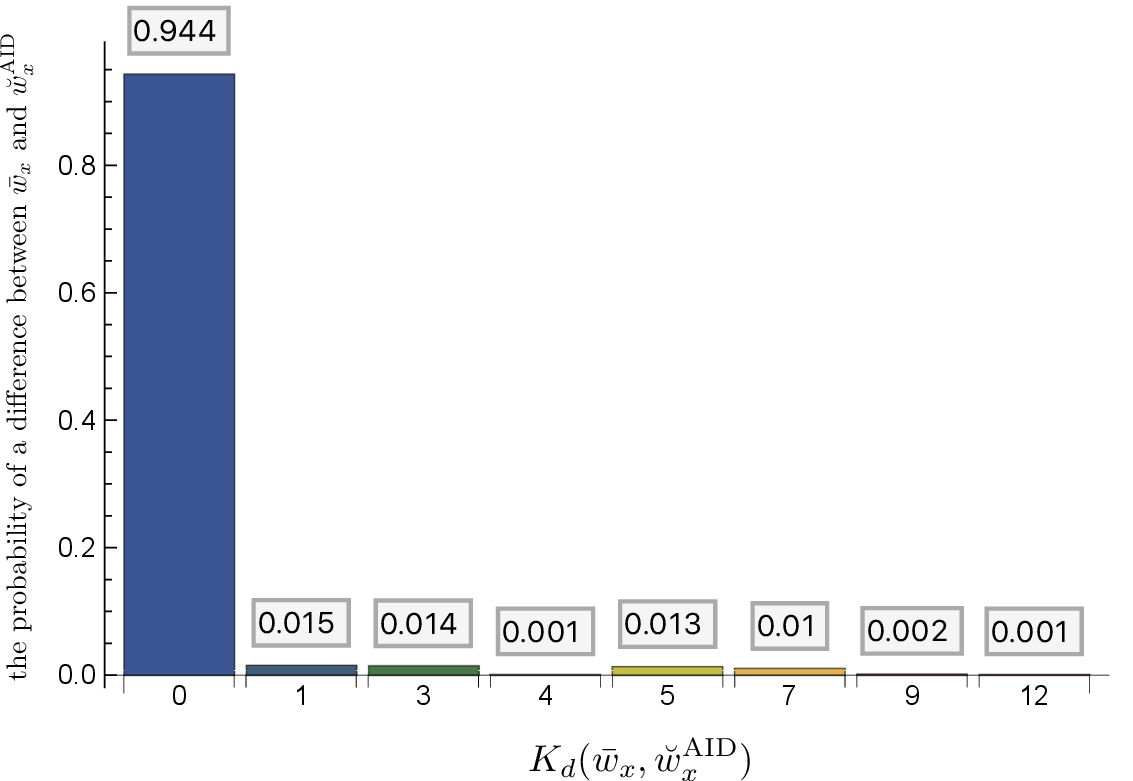}
\par\end{centering}
\caption{Estimated probability that for $I(S_{w_{x}\alpha_{k}})\protect\leq0.1$
the Kendall Tau distance $K_{d}(\bar{w}_{x},\breve{w}_{x}^{\text{AID}})$
will be: $0,1,\ldots,12$.}

\label{fig:fig_13_aid_kendall_prob}
\end{figure}

Similarly, for the AID grouping method, the chance that the ranking
will not change (i.e. $K_{d}(\bar{w}_{x},\breve{w}_{x}^{\text{AID}})=0$)
is $94.4\%$ (Fig. \ref{fig:fig_13_aid_kendall_prob}). The difference
of one transposition (i.e. $K_{d}(\bar{w}_{x},\breve{w}_{x}^{\text{AID}})=1$)
occurred in $1.5\%$ of cases, etc.

\begin{figure}[H]
\begin{centering}
\includegraphics[width=0.7\textwidth]{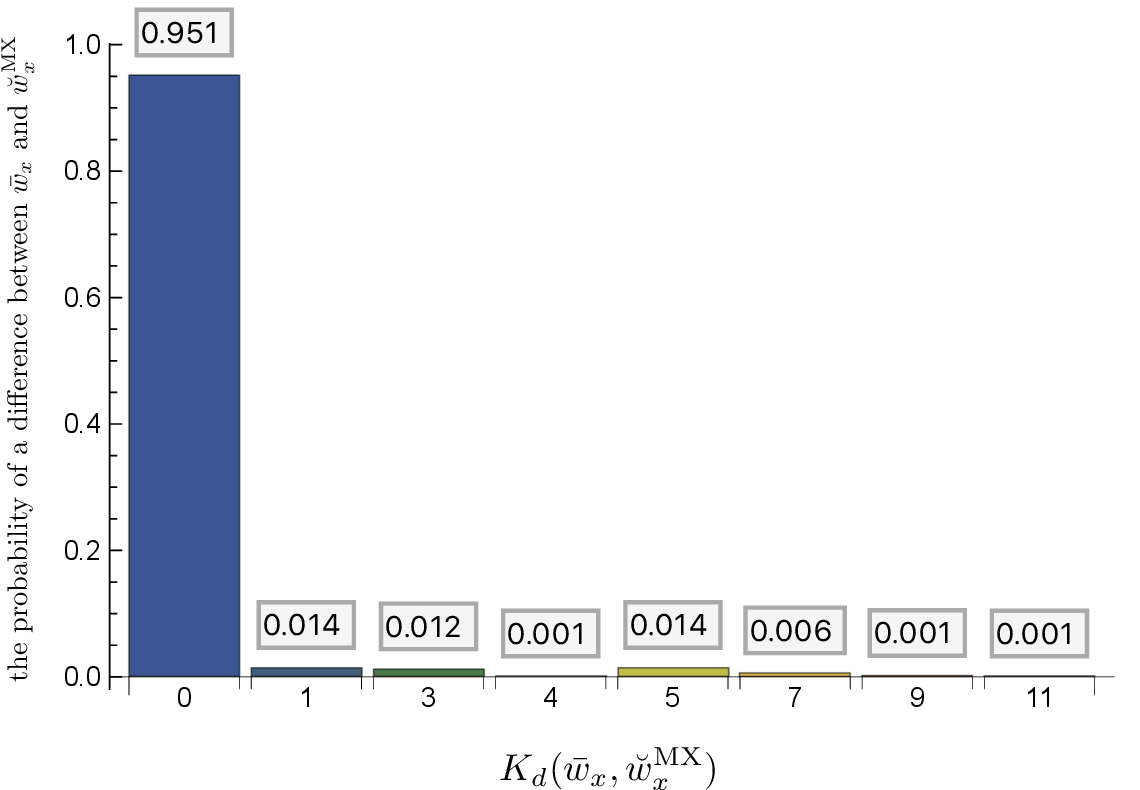}
\par\end{centering}
\caption{Estimated probability that for $I(S_{w_{x}\alpha_{k}})\protect\leq0.1$
the Kendall Tau distance $K_{d}(\bar{w}_{x},\breve{w}_{x}^{\text{MX}})$
will be: $0,1,\ldots,12$.}

\label{fig:fig_14_mx_kendall_prob}
\end{figure}

Finally, in the case of a mixed method the chance that the ranking
remains unchanged (i.e. $K_{d}(\bar{w}_{x},\breve{w}_{x}^{\text{MX}})=0$)
is $95.1\%$ (Fig. \ref{fig:fig_14_mx_kendall_prob}). Analogously,
the difference of one transposition (i.e. $K_{d}(\bar{w}_{x},\breve{w}_{x}^{\text{MX}})=1$)
occurred in $1.4\%$ of cases, three transpositions are needed to
transform $\bar{w}_{x}$ into $\breve{w}_{x}^{\text{MX}}$ in $1.2\%$
of cases and so on. 

\subsection{Experiments summary and discussion}

The conducted experiments used $4000$ sets containing $20$ matrices,
each simulating the opinions of experts in a group decision-making
process. The simulated scenarios contained from $5$ to $7$ alternatives.
The first of the Montecarlo experiments consisted in simulating a
simple manipulation and checking the robustness of the experts' ranking
aggregation. Three proposed methods were tested: APDD (Aggregation
of Preferential Distance-Driven expert prioritization), AID (Aggregation
of Inconsistency-Driven expert prioritization) and MX (the mixed method,
being a combination of the previous two). For all methods, their efficiency
depended on the average inconsistency of the matrix in the tested
20-element set. The higher the inconsistency, the lower the effectiveness.
In case of small inconsistencies (close to 0), the effectiveness of
the APDD method was close to $95\%$. i.e. in $95\%$ of cases out
of a hundred, this method was able to mitigate the negative effects
of the attack (Fig. \ref{fig:fig_3_apdd_success_ratio}). The AID
method fared slightly worse with effectiveness around $88\%$ (Fig.
\ref{fig:fig_4_aid_success_ratio}). The mixed approach seems to fall
between APDD and AID, with its effectiveness around $90\%$ (Fig.
\ref{fig:fig_5_mx_success_ratio}). For inconsistency around $0.1$,
the effectiveness of APDD drops to $89\%$ for WR (winner restoration)
and $86\%$ for RR (ranking restoration). The results for AID and
MX are respectively: $85\%$ (WR), $83\%$ (RR), $88\%$ (WR) and
$86\%$ (RR). Quantitative differences between the original (non-manipulated)
ranking and the ``fixed'' ranking were not large (Fig. \ref{fig:fig_6_apdd_distance},
\ref{fig:fig_7_aid_distance} and \ref{fig:fig_8_mx_distance}) and
varied from $0.0327$ to $0.047$. Interestingly, the mixed method
performed as well as the first APDD method.

The purpose of the second experiment was to test the proposed aggregation
methods on non-manipulated data. In this case, these methods seem
unnecessary and superfluous. Hence, changes in the differences between
the ranking obtained using these methods and the ranking obtained
using the classical method can be treated as an unnecessary disturbance.
We examined the generated data for quantitative and qualitative differences
between the two ranking vectors. We achieved the best quantitative
result for the MX method, with the average Manhattan distance of $0.009$
(\ref{eq:mx-quantitative-diff-2nd-mc-ex}). The AID method fared slightly
worse with the Manhattan distance $0.011$ (\ref{eq:aid-quantitative-diff-2nd-mc-ex}),
and at the end APDD with a result of $0.017$ (\ref{eq:apdd-quantitative-diff-2nd-mc-ex}).
In all of the cases, it can be seen that this distance depends on
the average inconsistency of experts and increases along with increasing
inconsistency.

To examine the qualitative difference between the classic aggregation
method and the modified methods, we used Kendall's tau distance measure
and the subset of data for which the average inconsistency is not
too high (less than $0.1$). With these assumptions, the MX method
turned out to be the best again, as in $95\%$ of cases the ranking
did not change (Fig. \ref{fig:fig_14_mx_kendall_prob}). The AID method
fared slightly worse since for 94\% of cases the order remained unchanged.
The APDD method took last position with $92\%$ untouched rankings.

The proposed methods are not perfect. However, almost $90\%$ of effectiveness
related to eliminating the negative effects of manipulation is achieved,
with a $5\text{\textendash}8\%$ risk of introducing disturbances
when the manipulation occurs. The negative impact in the second case
is much less severe when the ranking result is interpreted quantitatively,
i.e., when the rank value is more important than the position on the
list. Then, potentially significant changes introduced by the manipulation
can be elimiated or mitigated with relatively small quantitative changes
in the non-manipulated ranking. However, is it worth using modified
aggregation methods when the ranking result is ultimately given an
ordinal meaning? It depends on the subjective assessment of the people
responsible for the decision process. In other words, if the risk
of manipulation is not negligible, then it may be worth using the
presented aggregation methods. In the article, we proposed three heuristic
ranking aggregation methods, the third of which combined the other
two. Based on the conducted experiments, the third one is the most
effective in practice. However, this observation suggests that adding
more heuristics to identify possible manipulations could improve the
results. In particular, effective methods of mitigating the effects
of manipulation should simultaneously be based on many mutually complementary
approaches.

\section{Summary\label{sec:Summary}}

The article presents three modified procedures for aggregating expert
opinions that can be used in group decision making using the AHP method.
They allow for mitigation (or elimination) of the adverse effects
of manipulation with a small risk of distorting of the ranking. The
first two methods are based on heuristics that make the weight of
a given expert dependent on the level of its inconsistency and the
group's average opinion. The third method, perhaps the most promising,
combines the other two. Developing more secure and tamper-resistant
methods based on comparing alternatives in pairs will require further
study of attack and defense methods against manipulation. Thus, the
presented results bring us one step closer to this goal.

\section*{Acknowledgements }

The research was supported by the National Science Centre, Poland,
as a part of the project SODA no. 2021/41/B/HS4/03475 and by the Polish
Ministry of Science and Higher Education (task no. 11.11.420.004).

\bibliographystyle{elsarticle-harv}
\addcontentsline{toc}{section}{\refname}\bibliography{papers_biblio_reviewed}

\end{document}